\documentclass[journal]{IEEEtran}
\usepackage[utf8]{inputenc} 
\usepackage[T1]{fontenc}    
\usepackage{hyperref}       
\usepackage{url}            
\usepackage{booktabs}       
\usepackage{amsfonts}       
\usepackage{nicefrac}       
\usepackage{microtype}      
\usepackage{algorithm}
\usepackage{algorithmic}
\usepackage[export]{adjustbox}
\usepackage{bm}
\usepackage{amsmath,amssymb}
\usepackage{subfigure}
\usepackage{verbatim} 
\usepackage[super]{nth}
\usepackage{mathtools}
\usepackage{xcolor}
\DeclareMathOperator*{\E}{\mathbb{E}}
\DeclareMathOperator*{\argmin}{arg\,min}
\DeclarePairedDelimiter\floor{\lfloor}{\rfloor}
\ifCLASSOPTIONcompsoc

\usepackage[nocompress]{cite}
\else
\usepackage{cite}
\fi

\DeclareUnicodeCharacter{2212}{-}
\begin{document}

\title{Adaptive Experience Selection for Policy Gradient}
	
\author{Saad~Mohamad~and~Giovanni~Montana
		\IEEEcompsocitemizethanks{\IEEEcompsocthanksitem The authors are with Warwick Manufacturing Group (WMG), University of Warwick, United Kingdom. 
		E-mail:\{saad.mohamad, g.montana\}@warwick.ac.uk
		}}

\IEEEtitleabstractindextext{%
\begin{abstract}
Policy gradient reinforcement learning (RL) algorithms have achieved impressive performance in challenging learning tasks such as continuous control, but suffer from high sample complexity. Experience replay is a commonly used approach to improve sample efficiency, but gradient estimators using past trajectories typically have high variance. Existing  sampling strategies for experience replay like uniform sampling or prioritised experience replay do not explicitly try to control the variance of the gradient estimates. In this paper, we propose an online learning algorithm, adaptive experience selection (AES), to adaptively learn an experience sampling distribution that explicitly minimises this variance. Using a regret minimisation approach, AES iteratively updates the experience sampling distribution to match the performance of a competitor distribution assumed to have optimal variance. Sample non-stationarity is addressed by proposing a dynamic (i.e. time changing) competitor distribution for which a closed-form solution is proposed. We demonstrate that AES is a low-regret algorithm with reasonable sample complexity. Empirically, AES has been implemented for deep deterministic policy gradient and soft actor critic algorithms, and tested on 8 continuous control tasks from the OpenAI Gym library. Ours results show that AES leads to significantly improved performance compared to currently available experience sampling strategies for policy gradient.
\end{abstract}
}
\maketitle
\IEEEdisplaynontitleabstractindextext
\IEEEpeerreviewmaketitle
\section{Introduction}\label{sec:introduction}
Reinforcement learning (RL) is a computational approach for solving sequential decision-making problems under uncertainty~\cite{lin1993reinforcement}. In these problems, typically an agent interacts over time with the environment and learns to take actions according to an optimal policy that maximises the cumulative future expected rewards. Recent advances in RL have adopted deep neural networks as high-capacity function approximators resulting in deep reinforcement learning (DRL)~\cite{sutton2018reinforcement}. DRL has yielded impressive results in a number of tasks, including learning to play Atari games~\cite{mnih2015human}, controlling robots from raw images~\cite{levine2016end}, and mastering the game of Go~\cite{silver2016mastering}. 
	
Policy gradient algorithms~\cite{sutton2000policy,wang2016sample,de2018experience,lillicrap2015continuous,schulman2015trust,PPO_Schulman_2017} seek the optimal policy by operating directly on the gradient of accumulated rewards taken with respect to the policy parameters. These methods have reached excellent performance in problems with large and/or continuous action spaces~\cite{BenchmarkDRLContinuousControl_Duan_2016}. In its simplest formulation, the policy gradient is estimated from trajectories\footnote{A trajectory is a sequence of transitions, each including current state, action, next state and reward, up to a pre-defined time horizon.} generated by the current policy (i.e. on-policy)~\cite{williams1992simple,sutton2000policy,schulman2015trust,PPO_Schulman_2017}. On-policy gradient estimators are unbiased, but contemporary algorithms suffer from low sample efficiency. This is because a new set of trajectories need to be generated for each policy update, i.e. at every step.
	
To improve sample efficiency, experience replay (ER)~\cite{lin1992self} is commonly used. This approach works by storing trajectories generated by past policies and reusing them to estimate the policy gradient. ER leads to off-policy algorithms, i.e. the gradient of the current policy is estimated using trajectories generated by different policies~\cite{IW_Jie_2010,degris2012off,GuidedPolicySearch_Levine_2013,wang2016sample}. In these algorithms, the divergence between current and past policies leads to bias in gradient estimators, which is often corrected by employing an importance sampling ratio in the off-policy gradient estimator. However, this ratio is unbounded and can yield high (or even infinite) variance~\cite{IWVariance_Andradottir_1995,IWVariance_Precup_2001,IWVariance_Mahmood_2014,IWVariance_Schlegel_2019}. Such high variance compromises the algorithm's convergence, resulting in increased sample complexity and hindering effective learning. 
	
To reduce the variance in off-policy gradient estimators, several approaches have focused on seeking estimators with better bias-variance trade-offs ~\cite{degris2012off,precup2000eligibility,wang2016sample,Retrace_Munos_2016, IMPALA_Espeholt_2018,OffPolicyStateImportanceWeighting_Liu_2018,OffPolicyVarianceReductionControlPrior_Cheng_2019}. Most of these variance reduction algorithms uniformly sample trajectories from the replay butter to compute the gradients. Improved experience replay sampling methods have also been studied. A representative methodology is prioritised experience replay~\cite{schaul2015prioritized} which identifies the most important trajectories and sample them more frequently to improve learning efficiency whereas the the importance of a trajectory is determined by the temporal-difference error of transitions. Other experience replay sampling strategies have also been proposed such as distributed prioritized experience replay~\cite{horgan2018distributed}, and methods that depend on the propagation of sample priority~\cite{brittain2019prioritized}, curiosity~\cite{zhao2019curiosity}, the divergence between trajectories and current policy~\cite{ExperienceReplayMemoryForget_Novati_2019}, and a method to learn a separate policy for experience sampling~\cite{RLExperienceSampling_Zha_2019}. Although the available experience sampling methods can often achieve better performance compared to uniform sampling, none of them explicitly address the high variance issue.

In this paper, we learn a sampling distribution for selecting samples from the ER buffer to compute the gradient at each step. Unlike existing approaches, our aim is to adaptively choose this sampling distribution so that it explicitly minimises the variance of the gradient estimates. This is achieved through an online regret minimisation approach~\cite{cesa2006prediction}, which we call Adaptive Experience Selection (AES), whereby the sampling distribution is updated in a sequential manner during the learning phase. We assume that there exists an unknown competitor distribution that has optimal variance. AES attempts to update the current sampling distribution in order to match the competitor's variance. To address the non-stationarity of samples in the experience replay buffer, which is due to the process of sample insertion and over-writing, we consider the case of a dynamic (i.e. time-varying) competitor distribution. We demonstrate that AES leads to a closed-form solution of the sampling distribution, and has low regret and reasonable sampling complexity. Empirically, we have implemented AES with two representative DRL algorithms, deep deterministic policy gradient (DDPG)~\cite{lillicrap2015continuous} and soft actor critic (SAC)~\cite{SAC_Haarnoja_2018}, and have examined the performance on 8 continuous control tasks in OpenAI Gym library. We show that AES achieves significantly improved performance compared to existing experience sampling strategies.
\section{Background} \label{background}
	
\subsection{Markov Decision Processes}
	
We consider sequential decision making problems whereby an agent interacts with an environment, and the decision process is modelled as discrete-time Markov Decision Process (MDP). A MDP is defined by a tuple $M=\{\mathcal{S},\mathcal{A},P,r,\gamma,\rho_0\}$, where $\mathcal{S}$ is the state space; $\mathcal{A}$ is the action space; $P:\mathcal{S} \times \mathcal{A} \times \mathcal{S} \rightarrow \mathbb{R}$ is the transition probability distribution, $r: \mathcal{S} \times \mathcal{A} \rightarrow \mathbb{R}$ is the reward function, $\gamma \in \left(0,1\right)$ is the discounting factor and $\rho_0: \mathcal{S} \rightarrow \mathbb{R}$ the initial state distribution. At a timestep $t$, the agent observes the current state ${\bm s}_t \in \mathcal{S}$ and takes an action according to a policy ${\bm a}_t \sim \pi_{\bm \theta}\left({\bm a}_t|{\bm s}_t\right)$ with ${\bm \theta}$ denoting policy parameters. Then, the environment moves to the next state ${\bm s}_{t+1} \sim P\left({\bm s}_{t+1}|{\bm s}_t, {\bm a}_t\right)$, and the agent receives a reward $r\left({\bm s}_t,{\bm a}_t\right)$. With ${\bm s}_0 \sim \rho_0$, and after following a fixed policy $\pi_{\bm \theta}$ for $H$ steps, a trajectory $\tau = \left({\bm s}_0, {\bm a}_0, {\bm s}_1, {\bm a}_1, ... , {\bm s}_{H-1}, {\bm a}_{H-1}\right)$ is obtained. Let $R\left(\tau\right)=\sum_{t=0}^{H-1}\gamma^tr({\bm s}_t, {\bm a}_t)$ be the return for $\tau$. RL aims to maximise the expected return, denoted by $J$:
	\begin{equation}\label{main_eq}
	J({\bm \theta})=\E_{\tau\sim p(\tau|{\pi}_{\bm \theta})}R\left(\tau\right)
	\end{equation}
	where $p(\tau|{\pi}_{\bm \theta})$ is the trajectory distribution under policy $\pi_{\bm \theta}$, and is defined as: 
	\begin{equation}\label{eq:dist_tau}
	p(\tau|\pi_{\bm \theta})=\rho_0({\bm s}_0)\prod_{t=0}^{H-1}P({\bm s}_{t+1}|{\bm s}_{t}, {\bm a}_{t})\pi_{\theta}({\bm a}_t|{\bm s}_t)
	\end{equation}
	
\subsection{Policy gradient and experience replay} \label{experiencereplay}
	
	{\bf On-policy methods}. Policy gradient methods update $\bm \theta$ along the direction of $\nabla_{\bm \theta}J\left({\bm \theta}\right)$ to maximise $J\left({\bm \theta}\right)$. It can be shown that $\nabla_{\bm \theta}J\left({\bm \theta}\right)$ can be expressed as~\cite{sutton2000policy}:
	\begin{equation}\label{equu1}
	\nabla J({\bm \theta})=\E_{\tau\sim p(\tau|\pi_{\bm \theta})}\left[\nabla \log p(\tau|\pi_{\bm \theta})R(\tau)\right]
	\end{equation}
	The analytical expression of Eq.~\ref{equu1} is difficult to obtain, since environmental knowledge like transition probabilities and reward functions are difficult to obtain. Alternatively, Monte Carlo methods are widely used to estimate the expectation in Eq.~\ref{equu1} from trajectories only. The corresponding Monte Carlo estimator for Eq.~\ref{equu1} is:
	\begin{equation}\label{eq:pgmc}
	\hat{\nabla}J({\bm \theta})=\frac{1}{N}\sum_{k=1}^{N}\nabla \log p(\tau_k|\pi_{\bm \theta})R(\tau_k)
	\end{equation}
	where $\tau_{k}$ is the $k^{\mathrm{th}}$ trajectory in Monte Carlo sampling. $\hat{\nabla}J({\bm \theta})$ is an unbiased estimator of $\nabla J({\bm \theta})$, when $\tau_k$ is generated by $\pi_{\bm\theta}$. This can be shown by taking expectation with respect to $\tau_k\sim p(\tau_k|\pi_{\bm \theta})$ over the right hand side of Eq.~\ref{eq:pgmc}.
	
	On policy RL algorithms~\cite{williams1992simple,schulman2015trust,PPO_Schulman_2017} run the current policy to obtain $\tau_{k}$. After ${\bm \theta}$ is updated, a new $\tau_{i}$ is obtained under the new policy. This procedure requires to generate a large amount of new trajectories for each policy update, thus resulting in low sample efficiency. 
	
	{\bf Off-policy methods}. To increase sample efficiency, off-policy RL algorithms~\cite{IW_Jie_2010,degris2012off,GuidedPolicySearch_Levine_2013,wang2016sample} update the current policy using existing trajectories generated by previous, hence different, policies. Considering a target policy $\pi_{\bm \theta}$ to maximise $J\left({\bm \theta}\right)$ and a behaviour policy denoted by $\mu$ to generates trajectories for Monte Carlo gradient estimate,  Eq.~\ref{equu1} can be rewritten as:
	\begin{equation}\label{equpgop}
	\begin{aligned}
	\nabla J({\bm \theta})=&\E_{\tau \sim p(\tau|\mu)}\left[\omega\left(\tau|\pi_{\bm \theta}, \mu\right)\nabla \log p(\tau|\pi_{\bm \theta})R(\tau)\right]\\
	=&\E_{\tau \sim p(\tau|\mu)}\left[\omega\left(\tau|\pi_{\bm \theta}, \mu\right) g\left(\tau|\pi_{\bm \theta}\right)\right]
	\end{aligned}
	\end{equation}
	where $\omega\left(\tau|\pi_{\bm \theta}, \mu\right) = {p\left(\tau|\pi_{\bm \theta}\right)}/{p\left(\tau|\mu\right)}$ is the importance weight ratio, and $g\left(\tau|\pi_{\bm \theta}\right)=\nabla \log p(\tau|\pi_{\bm \theta})R(\tau)$. Note that Eq.~\ref{equpgop} samples $\tau$ from $\mu$ rather than $\pi_{\theta}$. This requires to introduce $\omega$ for the equivalence between Eq.~\ref{equpgop} and Eq.~\ref{equu1}. However, introducing $\omega$ results high variance. To see this, it suffices to note that, using Eq.~\ref{eq:dist_tau},
	\begin{equation}\label{eq:offpolicypgmciwr}
	\omega\left(\tau|\pi_{\bm \theta}, \mu\right) = \prod_{t=0}^{H-1}\frac{\pi_{\bm \theta}\left({\bm a}_{t}|{\bm s}_{t}\right)}{\mu\left({\bm a}_{t}|{\bm s}_{t}\right)}
	\end{equation}
	Eq.~\ref{eq:offpolicypgmciwr} is a product of many unbounded importance weight ratios. Therefore, if several $\mu\left({\bm a}_{t}|{\bm s}_{t}\right)$ have very low probabilities, the corresponding ratio $\pi_{\bm \theta}\left({\bm a}_{t}|{\bm s}_{t}\right)/\mu\left({\bm a}_{t}|{\bm s}_{t}\right)$ can become very large; the product of these ratio can explode resulting in very high $\omega$, and hence high variance. 
	
	Alternative formulations for $\omega$ and $g$ have been proposed to reduce the variance whilst keeping the bias low. For example, clipping~\cite{Retrace_Munos_2016,wang2016sample} or scaling~\cite{precup2000eligibility} the ratios $\pi_{\bm \theta}\left({\bm a}_{t}|{\bm s}_{t}\right)/\mu\left({\bm a}_{t}|{\bm s}_{t}\right)$ to prevent $\omega$ becoming too large; subtracting a baseline from $g$~\cite{wang2016sample,IMPALA_Espeholt_2018,OnOffPolicy_Fakoor_2019}; the baseline is designed to reduce the variance while adding little or no bias; taking the expectation in Eq.~\ref{equpgop} with respect to individual state-action pairs rather than trajectories to alleviate the issue of exploding importance ratios by considering a marginal return function with limiting state distribution~\cite{degris2012off,wang2016sample}.
	
	{\bf Experience replay}. Most off-policy RL algorithms use an experience replay buffer to store  the trajectories generated by the current policy at each update. The policy gradient can be estimated using Eq.~\ref{equpgop} from a batch of trajectories randomly sampled from the experience replay. Let $\mathcal{B}$ be the experience replay, and $\pi_{{\bm \theta}_t}$ be the policy parameters at timestep $t$. At a specific timestep $t$, the experience replay $\mathcal{B}$ may contain past trajectories $\pi_{{\bm \theta}_0}$, $\pi_{{\bm \theta}_1}$, ... $\pi_{{\bm \theta}_t}$. Assuming a uniform sampling distribution over $\mathcal{B}$, an unbiased Monte Carlo gradient estimator for Eq.~\ref{eq:offpolicypgmciwr} is:
	\begin{equation}\label{equpgopmc}
	\hat{\nabla}J({\bm \theta})=\frac{1}{|\Psi|}\sum_{k \in \Psi}\omega\left(\tau_{k}|\pi_{\bm \theta}, \pi_{{\bm \theta}_{l\left(k\right)}}\right) g\left(\tau_{k}|\pi_{\bm \theta}\right)
	\end{equation}
	where $\Psi$ is the set of sampled indexes for the trajectories in $\mathcal{B}$, and $l\left(k\right)$ is a function specifying the policy update step corresponding to the $k^{\mathrm{th}}$ sample in $\mathcal{B}$. 
\section{Methodology}
	
\subsection{Problem Formulation} \label{sec:methodology_problem_formulation}
	
	In this paper, we focus on off-policy policy gradient algorithms using experience replay. We seek a sampling distribution for the trajectories in the ER buffer such that the gradient estimator features the smallest possible variance whilst adding no bias. In this subsection, we present the proposed formulation.
	
	Let $U\left\{1,\mathcal{|B|}\right\}$ be a discrete uniform distribution as commonly adopted for experience sampling. We rewrite Eq.~\ref{equpgopmc} to incorporate the procedure of sampling experience replay into the policy gradient formulation:
	\begin{equation}\label{ee2_original_gradient}
	\nabla J\left({\bm \theta}\right)=\E_{k \sim U}{\E_{\tau \sim p(\tau|\pi_{{\bm \theta}_{l(k)}})}}{\omega\left(\tau|\pi_{\bm \theta}, \pi_{{\bm \theta}_{l\left(k\right)}}\right) g\left(\tau|\pi_{\bm \theta}\right)}
	\end{equation}
	where the outer expectation is taken with respect to the sampled index in $\mathcal{B}$, and the inner expectation is with respect to the selected trajectory. With Eq.~\ref{ee2_original_gradient}, using $|\Psi|$-sample Monte Carlo estimator for the outer expectation and single-sample Monte Carlo estimator for the inner expectation, leads to the estimator in Eq.~\ref{equpgopmc}. This procedure can be expressed as firstly selecting $\Psi$ then averaging the gradient over the selected trajectories.
	
	Instead of using uniform sampling, we aim to learn a sampling distribution that minimises the variance in the gradient estimate. Specifically, let ${\bm p}=[p(1), p(2), ... , p(|\mathcal{B}|)]$ be a vector, where $p(i)$ represents the sampling probability for the $i^{\mathrm{th}}$ trajectory in $\mathcal{B}$, and $\sum_{i=1}^{|\mathcal{B}|}{p(i)}=1$. Let $\mathcal{M}\left(1,{\bm p}\right)$ be a multinomial distribution parameterised by ${\bm p}$ with single trial. The policy gradient formulation with $\mathcal{M}$ as experience sampling distribution can be obtained by rewriting Eq.~\ref{ee2_original_gradient}:
	\begin{equation}\label{ee2_original_gradient_prob}
	\nabla J\left({\bm \theta}\right)=\E_{k \sim \mathcal{M}}{\lambda_k\E_{\tau}}{\: \omega\left(\tau|\pi_{\bm \theta}, \pi_{{\bm \theta}_{l\left(k\right)}}\right) g\left(\tau|\pi_{\bm \theta}\right)}
	\end{equation}
	where $\tau \sim p(\tau|\pi_{{\bm \theta}_{l(k)}})$, and the ratio $\lambda_k = 1/p\left(k\right)|\mathcal{B}|$ is the importance weight ratio. By using $|\Psi|$-sample Monte Carlo estimator for the outer expectation and single-sample Monte Carlo estimator for the inner expectation, we obtain a unbiased gradient estimator:
	\begin{equation}\label{ee2_original_gradient_prob_mc}
	\hat{\nabla}J({\bm \theta})=\frac{1}{|\Psi|}\sum_{k \in \Psi}\lambda_{k}\,\omega\left(\tau_{k}|\pi_{\bm \theta}, \pi_{{\bm \theta}_{l\left(k\right)}}\right) g\left(\tau_{k}|\pi_{\bm \theta}\right)
	\end{equation}
	We want to learn ${\bm p}$ so as to minimise the variance in each element of $\hat{\nabla}J({\bm \theta})$. Accordingly, we introduce the objective function:
	\begin{equation}\label{eq:obj0}
	f\left({\bm p}\right) = {\left \|\mathrm{Diag}\left[\mathrm{Cov}\left(\hat{\nabla}J({\bm \theta})\right)\right]\right \|}^{2}
	\end{equation}
	where $\mathrm{Cov}\left(\cdot\right)$ is the covariance matrix, $\mathrm{Diag}\left(\cdot\right)$ is an operator that extracts the diagonal elements in a square matrix and stacks them into a vector, and $\left\| \cdot \right\|$ is the Euclidean norm. The corresponding optimisation problem is:
	\begin{equation}\label{eq:obj1}
	\argmin_{{\bm p}\in\Delta}\:{f\left({\bm p}\right)}
	\end{equation}
	where $\Delta$ is the probability simplex.
	
	The objective function in Eq.~\ref{eq:obj0} can be expanded and simplified. Expanding $f\left({\bm p}\right)$ using the definition of covariance leads to:
	\begin{flalign}\label{eq:obj2}
	f\left({\bm p}\right) =& \E{{\left\| \hat{\nabla} J\left({\bm \theta}\right) - \E{\hat{\nabla} J\left({\bm \theta}\right)}  \right\|}^{2}} \nonumber\\
	=& \E {\left\| \hat{\nabla} J\left({\bm \theta}\right) \right\|}^2 - {\left\|\E {\hat{\nabla} J\left({\bm \theta}\right)}\right\|}^2
	\end{flalign}
	For brevity, we write $\omega\left(\tau_{i}|\pi_{{\bm \theta}}, \pi_{{\bm \theta}_{l\left(i\right)}}\right)$ as $\omega_{i}^{\bm \theta}$ and $g\left(\tau_{i}|\pi_{{\bm \theta}}\right)$ as $g_{i}^{\bm \theta}$. The formulation for $\E \,{\hat{\nabla} J\left({\bm \theta}\right)}$ is:
	\begin{flalign}\label{eq:obj3}
	\E \,{\hat{\nabla} J\left({\bm \theta}\right)} =&  \frac{1}{|\Psi|}\sum_{k \in \Psi}\E{\lambda_{k}\,\omega_{k}^{\bm \theta}\, g_{k}^{\bm \theta}} \nonumber\\
	=& \frac{1}{|\Psi|}\sum_{k \in \Psi}\sum_{i=1}^{|\mathcal{B}|}{p\left(i\right)\lambda_{i}\,\omega_{i}^{\bm \theta}\, g_{i}^{\bm \theta}} \nonumber\\
	=& \frac{1}{|\mathcal{B}|}\sum_{i=1}^{|\mathcal{B}|}{\omega_{i}^{\bm \theta}\, g_{i}^{\bm \theta}}
	\end{flalign}
	Thus, the second term in Eq.~\ref{eq:obj2} does not depend on ${\bm p}$, and can be ignored. On the other hand, let $B = \E{\sum_{i \in \Psi}\sum_{j \in \Psi,\,j \ne i}{\lambda_{i}\,\omega_{i}^{\bm \theta}\,{\left( g_{i}^{\bm \theta}\right)}^{\top}\,\lambda_{j}\,{\omega_{j}^{\bm \theta}\, g_{j}^{\bm \theta}}}} = \sum_{i \in \Psi}{\E{\lambda_{i}\,\omega_{i}^{\bm \theta}\,{\left( g_{i}^{\bm \theta}\right)}^{\top}\,\sum_{j \in \Psi,j \ne i}^{N}{\E{\lambda_{j}\,{\omega_{j}^{\bm \theta}\, g_{j}^{\bm \theta}}}}}}$. Assuming the Monte Carlo samples from $\mathcal{B}$ are i.i.d., the formulation for the first term in Eq.~\ref{eq:obj2} can be obtained:
	\begin{flalign}\label{eq:obj4}
	\E& {\left\| \hat{\nabla} J\left({\bm \theta}\right) \right\|}^2 = \frac{1}{{|\Psi|}^2}\E{ {\left\|\sum_{k \in \Psi}\lambda_{k}\,\omega_{k}^{\bm \theta}\, g_{k}^{\bm \theta}\right\|}^2} \nonumber\\
	=& \frac{1}{{|\Psi|}^2}\left(\sum_{k \in \Psi}\E{{\lambda_{k}^2\,{\left\|\omega_{k}^{\bm \theta}\, g_{k}^{\bm \theta}\right\|}_2^2}} + B\right) \nonumber\\
	=& \frac{1}{{|\Psi|}^2}\left( \sum_{k \in \Psi}\sum_{i=1}^{|\mathcal{B}|}{p\left(i\right)}{\lambda_{i}^2\,{\left\|\omega_{i}^{\bm \theta}\, g_{i}^{\bm \theta}\right\|}_2^2} + B\right) \nonumber\\
	=& \frac{1}{{|\Psi|}^2}\left( \sum_{i=1}^{|\mathcal{B}|}{\frac{|\Psi|}{p\left(i\right){|\mathcal{B}|}^2}}{{\left\|\omega_{i}^{\bm \theta}\, g_{i}^{\bm \theta}\right\|}_2^2} + B\right)
	\end{flalign}
	According to Eq.~\ref{eq:obj3}, the constant $B$ does not depend on ${\bm p}$. By substituting Eq.~\ref{eq:obj4} into Eq.~\ref{eq:obj2} and ignoring the constants $|\Psi|$, $|\mathcal{B}|$ and $B$, our objective function is simplified to:
	\begin{equation}\label{eq:obj5}
	f\left({\bm p}\right) = \sum_{i=1}^{|\mathcal{B}|}{\frac{1}{p\left(i\right)}}{{\left\|\omega_{i}^{\bm \theta}\, g_{i}^{\bm \theta}\right\|}_2^2}
	\end{equation}
	
	In the rest of this section, we present the proposed algorithm, Adaptive Experience Selection (AES) used to minimise Eq.~\ref{eq:obj5} in the context of off-policy RL. We start from the simplified setting where $\mathcal{B}$ is static, i.e. pre-filled, and the trajectories in $\mathcal{B}$ do not change. Then, we will consider the more general case where where sample insertions and overwriting are allowed during learning, as in general off-policy RL algorithms.
	
\subsection{Experience selection with static experience replay} \label{sec:AES_static_replay}
	
	In off-policy RL, the policy parameters are updated repeatedly using trajectories sampled from $\mathcal{B}$. In this setting, the optimisation in Eq.~\ref{eq:obj1} requires an online learning formulation whereby ${\bm p}$ is updated at each policy update step using the observed data from all the previous steps. Specifically, let ${\bm p}_t$ be the sampling distribution at a policy update step $t$. At a specific step $T$, we observe ${\left\{{\bm \theta}_t\right\}}_{t=1}^{T-1}$ and $\mathcal{B}$; we firstly obtain ${\bm p}_T$ by solving the following optimisation problem:
	\begin{equation}\label{eq:obj6}
	\begin{aligned}
	{\bm p}_T &\leftarrow \argmin_{{\bm p}\in\Delta}\:{\sum_{t=1}^{T-1}f_t\left({\bm p}\right)} \\
	\textrm{where} \quad &f_t\left({\bm p}\right) = \sum_{i=1}^{|\mathcal{B}|}{\frac{1}{{\bm p}\left(i\right)}}{{\left\|\omega_{i}^{{\bm \theta}_t}\, g_{i}^{{\bm \theta}_t}\right\|}_2^2}
	\end{aligned}
	\end{equation}
	Then, trajectories are sampled from $\mathcal{B}$ using ${\bm p}_T$ as sampling distribution; a gradient estimate is made using Eq.~\ref{ee2_original_gradient_prob}; the gradient estimate is used to update ${\bm \theta}_{T-1}$ to ${\bm \theta}_{T}$ with gradient descent. The above procedure is repeated for each policy update step.
	
	Optimisation algorithms assuming full data observation and i.i.d sampling like stochastic gradient descend are less applicable in this context for two main reasons. First, ${\bm \theta}_t$ is revealed sequentially for different $t$, hence we have incremental observations. Second, the ${\bm \theta}_t$'s are not i.i.d. as the policy parameters in subsequent steps depend on those in earlier steps. Here we resort to a regret minimisation approach to solve the above online learning problem. We define the regret at policy update step $T$, denoted by $f^R\left(T\right)$, as:
	\begin{equation}\label{eq:obj7}
	f^R\left(T\right) = \frac{1}{{|\mathcal{B}|}^2}{\left(\sum_{t=1}^{T}{f_t\left({\bm p}\right)}-\min_{{\bm p}\in\Delta}{\sum_{t=1}^{T}{f_t\left({\bm p}\right)}}\right)}
	\end{equation}
	where the first term in brackets is our objective function at step $T$, and the second term is a competitor assumed to have optimal ${\bm p}$. Ideally, we aim to update ${\bm p}$ to match the competitor's performance, formally $\lim_{T \rightarrow \infty}{\frac{1}{T}{f^R\left(T\right)}}=0$. This is also referred to as non-regret. 
	
	Follow-the-regularised-leader (FTRL) is an effective approach to solve the regret minimisation problem in Eq.~\ref{eq:obj7}, and can be formulated as~\cite{borsos2018online}:
	\begin{equation}\label{eq:ftrl}
	{\bm p}_T \leftarrow \argmin_{{\bm p}\in\Delta}{\sum_{t=1}^{T-1}{f_t\left({\bm p}\right)} + \nu\sum_{i=1}^{|\mathcal{B}|}\frac{1}{p\left(i\right)}}
	\end{equation}
	where the first term is our objective function; the second term is a regularisation term to avoid zero probability for any index; $\nu$ is a scalar balancing the two terms. Let $d_t\left(i\right)={\|\omega_i^{{\bm \theta}_t}g_i^{{\bm \theta}_t}\|}^2$. It can be shown that Eq.~\ref{eq:ftrl} has a closed-form solution~\cite{borsos2018online}:
	\begin{equation}\label{eq:ftrl_solution}
	p_T\left(i\right) = \frac{\sqrt{\sum_{t=1}^{T-1}{d_t\left(i\right)}+\nu}}{\sum_{i=1}^{|\mathcal{B}|}{\sqrt{\sum_{t=1}^{T-1}{d_t\left(i\right)}+\nu}}}.
	\end{equation}
	
	The update in Eq.~\ref{eq:ftrl_solution} leads to a regret bounded by $\mathcal{O}(\sqrt{T
	})$. To derive this, we need to make some assumptions on $\nabla J\left({\bm \theta}\right)$ and environmental rewards. In the rest of the paper we consider the general policy gradient formulation in Eq.~\ref{equpgop}. However, the assumptions and derivations below are also applicable to other policy gradient formulations with different forms of $\omega$ and $g$ as discussed in Section~\ref{experiencereplay}. We make three assumptions:\newline
	\textbf{Assumption 1 (Lower-bounded policy function)}: There exists a real constant $0<\beta\leq 1$, such that:
	\begin{equation}\label{eq:assum1}
	\nonumber\forall ({\bm s}_i, {\bm a}_i, {\bm \theta}_t) \quad \pi_{{\bm \theta}_t}({\bm a}_i|{\bm s}_i)\geq \beta
	\end{equation}
	\textbf{Assumption 2 (Lipschitz differential policy function)}: There exists a real constant $0 \leq L <\infty$, such that:
	\begin{equation}\label{eq:assum2}
	\nonumber\forall ({\bm s}_i, {\bm a}_i, {\bm \theta}_t) \quad ||\nabla \log \pi_{{\bm \theta}_t}({\bm a}_i|{\bm s}_i)|| \leq L
	\end{equation}
	\textbf{Assumption 3 (Bounded rewards)}: There exists a real constant $0 \leq \zeta < \infty$, such that: 
	\begin{equation}\label{eq:assum3}
	\nonumber\forall ({\bm s}_i, {\bm a}_i) \quad  \left| r \left({\bm s}_i,{\bm a}_i\right) \right| \leq \zeta
	\end{equation}
	Based on the above assumptions, a bound on $d_t$ is defined by the following lemma.\newline
	\textbf{Lemma 1}. Given Assumptions 1, 2 and 3, we have the following bound:
	\begin{equation}\label{eq:lemma1}
	\nonumber d_t\left(i\right) \leq {\left[\frac{\zeta(1-\gamma^{H})}{\beta^{H}(1-\gamma)}HL\right]}^2\hspace{-5mm}
	\end{equation}
	where $d_t\left(i\right)={\|\omega_i^{{\bm \theta}_t}g_i^{{\bm \theta}_t}\|}^2$. {\it Proof}. See Appendix~\ref{appendix:a_proofs_1}.\newline
	Given Lemma 1, the regret is bounded by the following corollary.\newline 
	\textbf{Corollary 1}. Given Assumptions 1, 2 and 3, we have the following regret bound:
	\begin{flalign}\label{eq:corollary1}
	\nonumber f^R\left(T\right) \leq \left(27\sqrt{T}+44\right){\left[\frac{\zeta(1-\gamma^{H})}{\beta^{H}(1-\gamma)}HL\right]}^2\hspace{-5mm}
	\end{flalign}
	{\it Proof}. The proof is straightforward by applying the above Lemma 1 with the Theorem 3 in~\cite{borsos2018online}.
\subsection{Experience selection with partial gradient} \label{sec:AES_partial_gradient}
	
	Eq.~\ref{eq:ftrl_solution} requires to compute the gradient for all the samples in $\mathcal{B}$ in all the policy update steps. Practically, this procedure is very computationally expensive, as $|\mathcal{B}|$ is usually in the order of millions. In this subsection, we aim to alleviate the computational load by using a subset of $\mathcal{B}$ to estimate $p_T$. This scheme fits well into an off-policy RL setting where, at each step, some trajectories are sampled from $\mathcal{B}$ to estimate the policy gradient; the same trajectories can then be used to estimate $p_T$. Let ${\Psi}_t$ be the index set of sampled trajectories in $\mathcal{B}$ at policy update step $t$. For $i \in {\Psi}_t$, an unbiased estimator of $d_t\left(i\right)$ is $d_t\left(i\right) / p_t\left(i\right)$, since $\E{[d_t\left(i\right) / p_t\left(i\right)]} = p_t\left(i\right)d_t\left(i\right) / p_t\left(i\right) = {d}_t\left(i\right)$. Therefore, we replace $d_t\left(i\right)$ in Eq.~\ref{eq:ftrl_solution} by
	\begin{equation}\label{eq:ftrl_solution_partial}
	\hat{d}_t\left(i\right) = 
	\begin{cases}
	\frac{d_t\left(i \right )}{p_t\left(i \right )} &\text{if } i \in \Psi_t\\
	0 \quad &\text{else}
	\end{cases}
	\end{equation}
	This leads to the following solution
	\begin{equation}\label{eq:ftrl_solution_partial_1}
	\hat{p}_T\left(i\right) = \frac{\sqrt{\sum_{t=1}^{T-1}{\hat{d}_t\left(i\right)}+\nu}}{\sum_{i=1}^{|\mathcal{B}|}{\sqrt{\sum_{t=1}^{T-1}{\hat{d}_t\left(i\right)}+\nu}}}
	\end{equation}
	One problem of Eq.~\ref{eq:ftrl_solution_partial_1} is that $\hat{d}_t\left(i\right)$ is unbounded for $i \in \Psi_t$. This leads to unbounded regret according to Lemma 1 and Corollary 1, as $d_t\left(i\right) / p_t\left(i\right)$ is unbounded. A typical solution is to mix Eq.~\ref{eq:ftrl_solution_partial_1} with the probability mass function of a uniform distribution:
	\begin{equation}\label{eq:ftrl_solution_partial_2}
	\tilde{p}_T\left(i\right) = (1-\kappa)\frac{\sqrt{\sum_{t=1}^{T-1}{\tilde{d}_t\left(i\right)}+\nu}}{\sum_{i=1}^{|\mathcal{B}|}{\sqrt{\sum_{t=1}^{T-1}{\tilde{d}_t\left(i\right)}+\nu}}} + \frac{\kappa}{|\mathcal{B}|}
	\end{equation}
	where $\kappa \in \left[0,1\right]$ is a coefficient to balance between the closed-form solution (the first term) and the uniform distribution (the second term). Since $\tilde{p}_T\left(i\right) \geq \kappa/|\mathcal{B}|$, we obtain a bound for $\tilde{d}_t\left(i\right)$ according to Lemma 1
	\begin{equation}\label{eq:ftrl_solution_partial_3}
	\tilde{d}_t\left(i\right)=\frac{d_t\left(i\right)}{\tilde{p}_t\left(i\right)} \leq \frac{|\mathcal{B}|}{\kappa}{\left[\frac{\zeta(1-\gamma^{H})}{\beta^{H}(1-\gamma)}HL\right]}^2
	\end{equation}
	Another effect of the uniform distribution in Eq.~\ref{eq:ftrl_solution_partial_2} is to improve exploration, as introducing uniform distribution encourages to sample each trajectory equally. In term of regret bound, we focus on expected regret as Eq.~\ref{eq:ftrl_solution_partial_2} is based on an estimator of $d_t(i)$. The following corollary demonstrates that Eq.~\ref{eq:ftrl_solution_partial_2} achieves a bound of $\mathcal{O}\left({|\mathcal{B}|}^{\frac{1}{3}}{T}^{\frac{2}{3}}\right)$ for expected regret.\newline
	\textbf{Corollary 2}. Let $\nu = {\left[\frac{\zeta(1-\gamma^{H})}{\beta^{H}(1-\gamma)}HL\right]}^2$ and $\kappa = {\left(|\mathcal{B}|/{T}\right)}^{1/3}$. Under Assumptions 1, 2 and 3, and assuming $T \geq |\mathcal{B}|$, Eq.~\ref{eq:ftrl_solution_partial_2} leads to the following bound:
	\begin{equation}\label{eq:ftrl_solution_partial_5}
	\begin{aligned}
	\nonumber\frac{1}{{|\mathcal{B}|}^2}\E&{\left(\sum_{t=1}^{T}{f_t\left(\tilde{\bm p}_t\right)}-\min_{{\bm p}\in\Delta}{\sum_{t=1}^{T}{f_t\left({\bm p}\right)}}\right)} \\
	&\leq 74{\left[\frac{\zeta(1-\gamma^{H})}{\beta^{H}(1-\gamma)}HL\right]}^2{|\mathcal{B}|}^{\frac{1}{3}}T^{\frac{2}{3}}
	\end{aligned}
	\end{equation}
	{\it Proof}. Note that the optimal ${\bm p}$ does not depend on $\tilde{\bm p}_t$. Thus, the proof can be done by applying the Theorem 7 in~\cite{borsos2018online} with Eq.~\ref{eq:ftrl_solution_partial_3}.
\subsection{Naive adaptive experience selection} \label{sec:AES_naive}
	
	The methods described in Section~\ref{sec:AES_static_replay} and Section~\ref{sec:AES_partial_gradient} assumes the experience replay is static, i.e. the experience replay is pre-filled and does not change over time. In off-policy RL, however, the experience replay is dynamically updated with new trajectories being inserted and old trajectories overwritten at every step. A naive application of the solution in Eq.~\ref{eq:ftrl_solution_partial_2} would reinitialise the sampling distribution each time new experiences are added into the buffer. Alg.~\ref{alg1_new} is the corresponding algorithm encapsulating this approach. Line 3 -- 11 generate trajectories using the current policy and store the trajectories into $\mathcal{B}$. Line 12 resets $\omega(i)$ which equals to $\sum_{t=1}^{T-1}{\tilde{d}_t\left(i\right)}$ in Eq.~\ref{eq:ftrl_solution_partial_2}. Line 14 updates the sampling distribution (analogous to Eq.~\ref{eq:ftrl_solution_partial_2}). Line 15 samples from experience replay with the current sampling distribution and update policy parameters. Line 16 updates $\omega(i)$ in an incrementally.
	
	A significant problem of Alg.~\ref{alg1_new} is that it is sample inefficient with regards to variance reduction. That is, it requires a significantly large number of samples to achieves low regret, as seen with the following results.\newline
	\textbf{Assumption 4 (Limited off-policy iterations)}. With Alg.~\ref{alg1_new}, we have $m<|\mathcal{B}|/|\Psi|$.\newline
	Assumption 4 results $m|\Psi|<|\mathcal{B}|$, i.e. we do not go through all the trajectories in $|\mathcal{B}|$ in a single epoch. This is reasonable in the RL context as doing so harms exploration and is very likely to get stuck into local optimum.\newline
	\textbf{Corollary 3}. Under assumption 4 and by setting $m=E/C^2$ implying that $T<C^2\frac{|\mathcal{B}|}{|\Psi|}$, for any constant $C>0$, Alg.~\ref{alg1_new} achieves the following bound:
	\begin{align}\label{eq:Corollary 3}
    \nonumber\frac{1}{E}{\sum_{k=1}^{E}{\frac{1}{m}\E{\frac{1}{{|\mathcal{B}|}^2}\left(\sum_{t=1}^{m}{f_t\left(\tilde{\bm p}_t\right)}-\min_{{\bm p}\in\Delta}{\sum_{t=1}^{m}{f_t\left({\bm p}\right)}}\right)}}}\\\nonumber\leq\mathcal{O}(\frac{|\mathcal{B}|^{1/3}C^{2/3}}{E^{1/3}})
	\end{align}
	{\it Proof}. See Appendix.~\ref{appendix:a_proofs_corollary_3}.\newline
	Corollary 3 demonstrates that for  Alg.~\ref{alg1_new} to achieve regret bound less than $\epsilon$, it requires number of samples $nb>H\frac{|\mathcal{B}|C^2}{\epsilon^3}$. Moreover, the condition $T<C^2\frac{|\mathcal{B}|}{|\Psi|}$ restricts the number of iterations allowed for the bound to hold. Primarily to achieve certain regret bound the number of iterations needs to be high. For higher $T$, we need to increase $C$ meaning that $m=\frac{{T}}{C^2}$ goes down resulting in less RL updates and lower RL convergence rate. 

	
	\begin{algorithm}[t]
		\caption{Naive Adaptive Experience Selection}\label{alg1_new}
		\begin{algorithmic}[1]
			\STATE \textbf{Input}: Number of epoch $E$, episode length $H$, number of off-policy iterations $m$, experience replay $\mathcal{B}$, batch size ${|\Psi|}$, parameters $\nu$ and $\kappa$
			\STATE \textbf{Initialise}: Current policy parameters ${\bm \theta}^{\prime}$, $\mathcal{B}=\varnothing$
			\STATE \textbf{Warm-up}: Obtaining some trajectories by running $\pi_{\theta^{\prime}}$ and add them to $\mathcal{B}$
			\FOR{$k=1,...,E$} 
			\STATE Set $t\leftarrow 0$ and get state $s_0$
			\WHILE{$t<H$}
			\STATE Perform ${\bm a}_t$ according to $\pi_{{\bm\theta}^{\prime}}(.|{\bm s}_t)$
			\STATE Get reward $r_t$ and next state ${\bm s}_{t+1}$
			\STATE Add experience (${\bm s}_t,{\bm a}_t,\pi_{{\bm\theta}^{\prime}}({\bm a}_t|{\bm s}_t), r_t, {\bm s}_{t+1})$ into $\mathcal{B}$
			\STATE $t\leftarrow t+1$
			\ENDWHILE
			\STATE Reset ${w(i)=0}$ for $i \in [1, 2, ..., |\mathcal{B}|]$
			\FOR{$t=1, 2, ... , m$}
			\STATE Update sampling distribution $ p_t(i)=(1-\kappa)\frac{\sqrt{ w(i)+\nu}}{\sum_{i=1}^{|\mathcal{B}|}\sqrt{w(i)+\nu}}+\kappa/{|\mathcal{B}|}$
			\STATE Sample index set ${\Psi}_t$ from $\mathcal{B}$ using ${\bm p}_t$ as sampling distribution; use the corresponding trajectories in $\mathcal{B}$ to update the policy parameters from ${\bm \theta}^{\prime}$ to ${\bm \theta}_{t}$ using off-policy policy gradient algorithms like DDPG or SAC
			\STATE Use the computed gradients in the step 15 to compute $\tilde{d}_{t}(i)$ and update $w(i)\leftarrow w(i)+\tilde{d}_{t}(i)$ for ${i\in \Psi_t}$ 
			\STATE Update current policy ${\bm \theta}^{\prime} \leftarrow {\bm \theta}_t$
			\ENDFOR
			\ENDFOR
		\end{algorithmic}
\end{algorithm}	
\subsection{Non-regret adaptive experience selection} \label{sec:AES_dynamic_regret}
In this subsection, we extend the methods described in Section~\ref{sec:AES_static_replay} and Section~\ref{sec:AES_partial_gradient} to general experience replay, i.e. when new experiences can be added on-the-fly and old experiences can be overwritten. This extension leads to a low-regret experience selection algorithm. In this more general setting, the optimal sampling distribution changes continuously as new experiences are being added into the buffer. Accordingly, we consider a regret minimisation model with a dynamic competitor formulated as
\begin{equation}\label{eq:obj_dyn_regret}
f^R\left(T\right) = \frac{1}{{|\mathcal{B}|}^2}{\left(\sum_{t=1}^{T}{f_t\left({\bm p}\right)}-{\sum_{t=1}^{T}{\min_{{\bm p}_t \in \Delta}{f_t\left({\bm p}_t\right)}}}\right)}
\end{equation}
The second term allows the competitor to choose different optimal sampling distribution for each policy update steps. To derive a solution, we make the following assumption.
\newline
\textbf{Assumption 5 (Lipschitz continuous gradient)}. There exists a real constant $0 < K<\infty$, such that:
\begin{equation}\label{eq:assum5}
\nonumber||g_{i}^{\bm \theta}-g_{i}^{\bm \theta'}||\leq K||\bm \theta-\bm \theta'||
\end{equation}
This is a mild smoothness assumption that holds for most models, particularly neural networks.\newline
\textbf{Assumption 6 (PL inequality)}. There exists a real constant $\xi>0$, such that:
\begin{equation}\label{eq:assum6}
\nonumber J\left({\bm \theta}\right)-J\left({\bm \theta^*}\right)\leq (2\xi)^{-1}||\nabla J\left({\bm \theta}\right)||^2
\end{equation}
where ${\bm\theta}^*=\argmin_{\bm\theta}J\left({\bm \theta}\right)$. This PL inequality assumption~\cite{nesterov2013gradient} is a bit stronger than the smoothness assumption. Examples of functions satisfying PL condition include neural networks with
one-hidden layers, ResNets with linear activation and objective functions in matrix factorisation~\cite{foster2018uniform}. 

It follows that we can bound the dynamic regret in Eq.~\eqref{eq:obj_dyn_regret} given that the change in $f_t\left({\bm p}\right)$ is small for consecutive steps. This smooth change can be guaranteed if only a very small portion of $\mathcal{B}$ is changed between two consecutive steps, which is common for off-policy RL algorithms. We propose Adaptive Experience Selection (AES) algorithm that exploits this smooth changing properties by regularly forgetting the influence of $f_t\left({\bm p}\right)$ for old $t$. AES updates ${\bm p}$ using FTRL and reset ${\bm p}$ every $M$ steps. This is formally formulated as
\begin{equation}\label{eq:ftrl_dyn}
{\bm p}_T \leftarrow \argmin_{{\bm p}\in\Delta}{\sum_{t=T_0}^{T-1}{f_t\left({\bm p}\right)} + \nu\sum_{i=1}^{|\mathcal{B}|}\frac{1}{{\bm p}\left(i\right)}}
\end{equation}
where $T_0 = \max(1,\floor{T/M}M)$ is the starting index of the $M$-step interval that $T$ falls in, with $\floor{\cdot}$ the floor function. Eq.~\ref{eq:ftrl_dyn} has a closed-form solution similarly to Eq.~\ref{eq:ftrl_solution}:
\begin{equation}\label{eq:ftrl_dyn_solution}
p_T\left(i\right) = \frac{\sqrt{\sum_{t=T_0}^{T-1}{d_t\left(i\right)}+\nu}}{\sum_{i=1}^{|\mathcal{B}|}{\sqrt{\sum_{t=T_0}^{T-1}{d_t\left(i\right)}+\nu}}}
\end{equation}
Equation~\ref{eq:ftrl_dyn_solution} is computationally expensive, since this formulation requires to calculate the gradient for all the trajectories in $\mathcal{B}$. Similarly to Section~\ref{sec:AES_partial_gradient}, we alleviate the computational load by making an unbiased estimate of $d_t$ using the sampled trajectories in each policy update step. This leads to the following update similarly to Eq.~\ref{eq:ftrl_solution_partial_2}:
\begin{equation}\label{eq:ftrl_dyn_solution_partial}
\tilde{p}_T\left(i\right) = (1-\kappa)\frac{\sqrt{\sum_{t=T_0}^{T-1}{\tilde{d}_t\left(i\right)}+\nu}}{\sum_{i=1}^{|\mathcal{B}|}{\sqrt{\sum_{t=T_0}^{T-1}{\tilde{d}_t\left(i\right)}+\nu}}} + \frac{\kappa}{|\mathcal{B}|}
\end{equation}
The expected regret with Eq.~\ref{eq:ftrl_dyn_solution_partial} is bounded by the following Theorem.\newline
\textbf{Theorem 1}. Under assumption 1, 2, 3, 5, 6. Given a condition on the learning rate according to the smoothness degree of $g_{i}^{\bm \theta}$: $\alpha_t<\frac{1}{K}$ and a condition on the severity of  overwriting  $M^2<|\mathcal{B}|$. Eq.~\ref{eq:ftrl_dyn_solution_partial} leads to the following bound:
\begin{equation}\label{eq:ftrl_dyn_solution_partial_2}
\begin{aligned}
    \nonumber &\frac{1}{{|\mathcal{B}|}^2}\E\bigg(\sum_{t=1}^{T}{f_t(\tilde{\bm p}_t)}-{\sum_{t=1}^{T}{\min_{{\bm p} \in \Delta}{f_t({\bm p}))}}\bigg)}\leq  \mathcal{O}(\frac{\kappa |\mathcal{B}|T}{M})\\&+\mathcal{O}(\frac{T}{M\sqrt{\kappa M}})+\mathcal{O}(\frac{|\mathcal{B}|T}{\kappa M})+\mathcal{O}(\frac{T}{\sqrt{|\mathcal{B}|M}})
\end{aligned}
\end{equation}
According to Theorem 1, setting $M=\frac{T}{C}$ leads to sub-linear regret bound with respect to $T$:
\begin{equation}\label{sbureg}
\begin{aligned}
&\frac{1}{{|\mathcal{B}|}^2}\E\bigg(\sum_{t=1}^{T}{f_t(\tilde{\bm p}_t)}-{\sum_{t=1}^{T}{\min_{{\bm p} \in \Delta}{f_t({\bm p}))}}\bigg)}\leq  \mathcal{O}(\kappa |\mathcal{B}|C)\\&+ \mathcal{O}(\frac{C^{\frac{3}{2}}}{\sqrt{\kappa T}})+\mathcal{O}(\frac{|\mathcal{B}|C}{\kappa })+\mathcal{O}(\frac{\sqrt{CT}}{\sqrt{|\mathcal{B}|}})
\end{aligned}
\end{equation}

However, the condition $M^2<|\mathcal{B}|$ implies $T<C\sqrt{{|\mathcal{B}|}}$ which restricts the number of iterations allowed for the bound to hold. Hence, there is a trade-off between the bound that can be achieved and the number of iteration $T$ allowed to achieve that bound. Note that primarily to achieve certain regret bound the number of iterations needs to be high. Thus, to relax the restriction on $T$, we set $M=\frac{\sqrt{T}}{C}$ so that $T<C^2{|\mathcal{B}|}$. Hence, we have:
\begin{equation}\label{sbureg2}
\begin{aligned}
    \nonumber &\frac{1}{{|\mathcal{B}|}^2}\E\bigg(\sum_{t=1}^{T}{f_t(\tilde{\bm p}_t)}-{\sum_{t=1}^{T}{\min_{{\bm p} \in \Delta}{f_t({\bm p}))}}\bigg)}\leq  \mathcal{O}({C\kappa |\mathcal{B}|\sqrt{T}})\\&+\mathcal{O}(\frac{CT^{1/4}}{\sqrt{\kappa}})+\mathcal{O}(\frac{C|\mathcal{B}|\sqrt{T}}{\kappa})+\mathcal{O}(\frac{C T^{3/4}}{\sqrt{|\mathcal{B}|}})
\end{aligned}
\end{equation}
This bound is softer that the one in Eq.~\eqref{sbureg}. However, the restriction on the number of iteration is more relaxed $T<C^2{|\mathcal{B}|}$. Note that to achieve this bound, we need high $T$. For higher $T$, we need to increase $C$ meaning that $M=\frac{\sqrt{T}}{C}$ goes down, hence more resetting is required. 

Similar study is applied to Alg.~\ref{alg1_new} in Corollary 3 which shows that to achieve regret bound less than $\epsilon$, Alg.~\ref{alg1_new} requires number of samples $nb>H\frac{|\mathcal{B}|C^2}{\epsilon^3}$ while AES requires $nb>H\frac{C^2C^{2/3}}{\epsilon^{4/3}}$. As stated above, for higher $T$, we need to increase $C$ meaning that $m=\frac{{T}}{C^2}$ goes down faster for Alg.~\ref{alg1_new} compared to $M$ of AES. Unlike AES, lower $m$ for Alg.~\ref{alg1_new} means less RL update and lower RL convergence rate. Finally, we should point out that the regret bound of AES is, unlike Alg.~\ref{alg1_new}, with respect to Dynamic competitor. 
\begin{algorithm}[t]
   \caption{Non-Regret Adaptive Experience Selection }\label{alg2}
\begin{algorithmic}[1]
\STATE \textbf{Input:} Number of iterations $T$, episode length $H$, restarting period $M$, experience size $|\mathcal{B}|$, sampling batch $|\Psi|$, parameters $\nu$ and $\kappa$
\STATE \textbf{Initialise:} Current policy parameters ${\theta_0}$, $\mathcal{B}=\varnothing$, $w(i)=0$ for $i \in [1, 2, ... , |\mathcal{B}|]$
\STATE \textbf{Warm-up}: Obtaining some trajectories by running $\pi_{\theta_0}$ and add them to $\mathcal{B}$
       \FOR{$t=1,2,...T$}
       \STATE Update sampling distribution $ p_t(i)=(1-\kappa)\frac{\sqrt{ w(i)+\nu}}{\sum_{i=1}^{|\mathcal{B}|}\sqrt{w(i)+\nu}}+\kappa/{|\mathcal{B}|}$
       \STATE Sample index set ${\Psi}_t$ from $\mathcal{B}$ using ${\bm p}_t$ as sampling distribution; use the corresponding samples in $\mathcal{B}$ to update the policy parameters from ${\bm\theta}^{\prime}$ to ${\bm\theta}_{t}$ using policy gradient algorithms like DDPG or SAC
       \STATE Use the computed gradients in the step 6 to compute $\tilde{d}_{t}(i)$ and update $w(i)\leftarrow w(i)+\tilde{d}_{t}(i)$ for ${i\in \Psi_t}$
       \IF{$t\%M==0$}
       \STATE Reset ${w(i)=0}$ for $i \in [1, 2, ..., |\mathcal{B}|]$
       \ENDIF
       \STATE Set $k \leftarrow 0$ and get state $s_0$
       \WHILE{$k<H$}
       \STATE Perform $a_k$ according to $\pi_{{\bm \theta}_t}(.|{\bm s}_k)$
       \STATE Get reward $r_k$ and next state ${\bm s}_{k+1}$
       \STATE Add experience (${\bm s}_k,{\bm a}_k,\pi_{{\bm\theta}_t}({\bm a}_k|{\bm s}_k), r_k, {\bm s}_{k+1})$ into $\mathcal{B}$; overwritten the experience sampled with distribution $1-{\bm p}_t$ if $\mathcal{B}$ is full
       \STATE $k\leftarrow k+1$
       \ENDWHILE
     \ENDFOR
\end{algorithmic}
\end{algorithm}

Based on the above studies, we propose sample efficient adaptive experience selection algorithm for off-policy policy gradient methods. The proposed algorithm is presented in Alg.~\ref{alg2}. Line 5 calculates the sampling distribution. Line 6 samples from experience replay with the current sampling distribution and update policy parameters. Line 7 incrementally update $w(i)$ for the sampling distribution estimation in the next iteration. Line 8--10 reset the sampling distribution every $M$ iterations. Line 12--17 generate experiences with the current policy and add them into the experience replay. Note that line 15 use $1-{\bm p}_t$ to sample the experience to be written by the new one. This is because we consider the experience with smaller sampling probability less valuable.
\section{Related work}
{\bf Variance reduction in policy gradient RL}. Control variate~\cite{BookControlVariate_Ross_2006} is typically used to reduce variance by subtracting a baseline from gradient estimate whilst adding no or little bias. Various forms of baselines has been exploited in RL, e.g. those based on exponential moving averages of rewards~\cite{ThesisRL_Sutton_1984,williams1992simple}, a closed-form formulation minimising the variance for each element in gradient~\cite{peters2008reinforcement}, an approximation of value function~\cite{schulman2015trust,A3C_Mnih_2016,wang2016sample,PPO_Schulman_2017}, a function based on past gradients~\cite{papini2018stochastic}, and a first order Taylor expansion of value function using off-policy data~\cite{OnOffPoliceQProp_Gu_2017}. Other representative methods include trust region regularisation that limits the policy change in each policy update~\cite{schulman2015trust,wang2016sample}, clipping or scaling the importance sampling ratio for importance sampling-based methods~\cite{precup2000eligibility,Retrace_Munos_2016,wang2016sample}, formulating policy gradient using limiting state distribution~\cite{degris2012off,wang2016sample}, and combining on-policy and off-policy methods~\cite{OnOffPoliceQProp_Gu_2017,OnOffPoliceQProp_Gu_2017,OnOffPolicyInterpolation_Gu_2017}. All these methods focus on seeking a low-variance estimation of the gradient, using either uniform sampling or temporal difference error-based sampling for trajectories. None of these works has investigated the problem of learning an adaptive sampling distribution to reduce variance.

{\bf Experience selection in RL}. A substantial body of work has also been devoted to design better sampling distributions for the ER buffer.~\cite{schaul2015prioritized} propose prioritised experience replay that uses a non-uniform sampling distribution prioritising the experiences with higher temporal-difference error.~\cite{brittain2019prioritized} propagate the priorities of samples through sequence of transitions.~\cite{horgan2018distributed} propose a distribution version of prioritised experience replay using multiple workers to generate and select experiences.~\cite{zhao2019curiosity} propose a curiosity-based strategy that prioritises samples with rarely-seen states.~\cite{ExperienceReplayMemoryForget_Novati_2019} consider the ``off-policyness'' of trajectories and ignore the trajectories that deviate too much from the current policy.~\cite{RLExperienceSampling_Zha_2019} learn a policy through RL to sample from experience replay. The influence of the size of experience replay~\cite{ExperienceReplaySize_Liu_2017, ExperienceReplaySize_Zhang_2017, de2018experience} and overwritten strategy of existing experiences~\cite{de2018experience} have also been studied. All the above methods do not explicitly optimise the sampling distribution to reduce the gradient's variance.
\section{Experiments}\label{sec6}
\subsection{Experiment setting}
{\bf Environments}. We perform experiments on 8 Mujoco~\cite{todorov2012mujoco} environments using the OpenAI Gym library~\cite{brockman2016openai}: InvertedPendulum, InvertedDoublePendulum, Reacher, Hopper, HalfCheeta, Walker2d, Ant, Humanoid. The environments are selected to include tasks with varying complexity. Examples for these environments are presented with some comparisons on our team YouTube channel~\footnote{https://www.youtube.com/channel/UCkDyucGZSYrSbBefntPZVHg}

{\bf Parameter Tuning}. All experiments are evaluated using 5 different seeds: $\{2,20,200,2000,20000\}$. For AES's hyper-parameters tuning, we set the seed to $2$, then use the hyper-parameters with the best results on the remaining 4 seeds. Details about the hyper-parameters setting and implementations of AES and the policy gradient RL methods (i.e., SAC and DDPG) can be found in App.~\ref{impl}. 

{\bf Evaluation}. We report plots (Fig.~\ref{fi4} and Fig.~\ref{fi5}) of the mean and standard deviation of the episodic return over all 5 different seeds. For each seed, the episodic return is computed on testing trials over the learning steps. This allows us to study the learning progress with respect to the number of samples acquired over time. AES' goal is to improve the sample efficiency by explicitly reducing the variance in gradient estimators. Lower variance should also entail improvement in the learning stability, robustness and final performance. We analyse these performance metrics from the reported plots. Performance measurements reflecting these four aspects are extracted from the plots and reported in App.~\ref{nummea}. A simple study of the variance over learning steps is reported in App.~\ref{variastud}.

{\bf Comparison methods}. We have implemented the proposed AES methodology for two widely used off-policy RL algorithms: deep deterministic policy gradient (DDPG)~\cite{lillicrap2015continuous} (see App.~\ref{sec5}) and soft actor-critic (SAC)~\cite{SAC_Haarnoja_2018} (See App.~\ref{sec4}), referred to as AES-DDPG and AES-SAC, respectively. We compare these algorithms with two competing methods: DDPG and SAC with uniform experience sampling as baseline and DDPG and SAC with prioritised experience replay~\cite{schaul2015prioritized} (here called pri-DDPG and pri-SAC, respectively) as representative methods for experience selection.
\subsection{Results}
\begin{figure*}[ht]
\subfigure[Inverted Pendulum]{\includegraphics[width = 0.33\textwidth, height=0.27\textwidth]{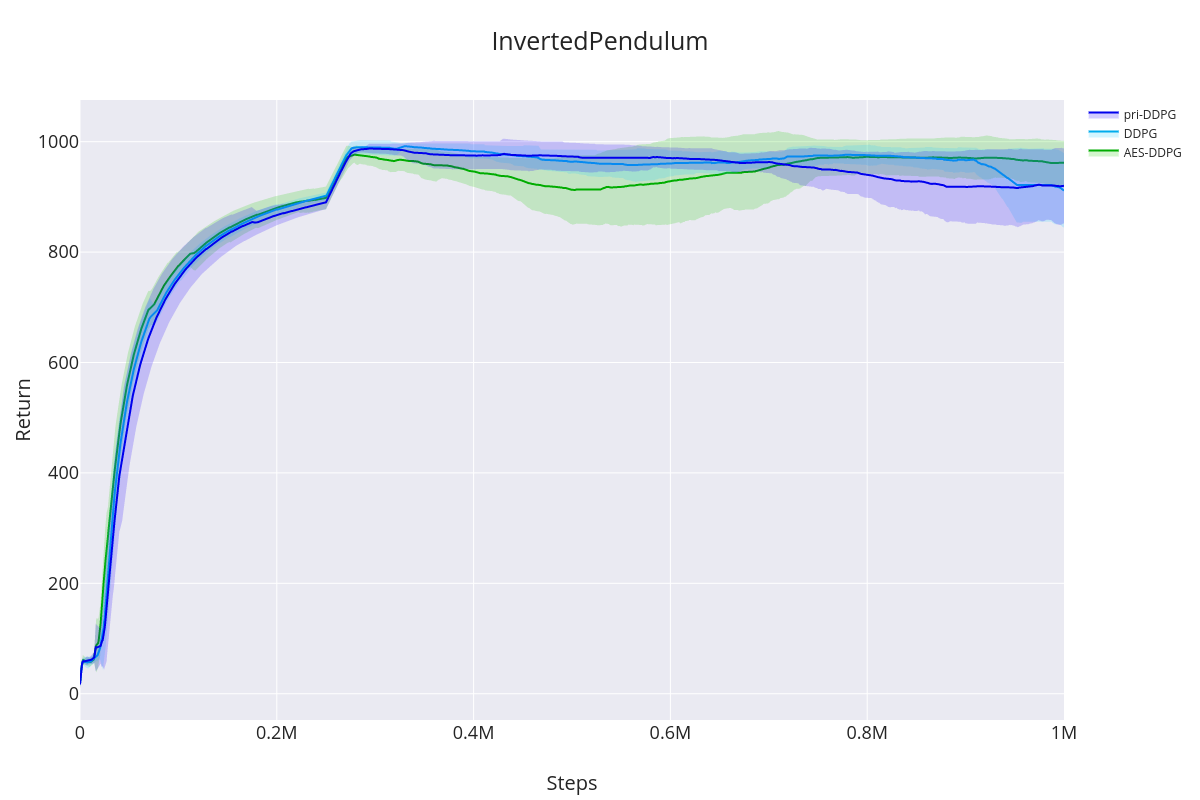}}
\subfigure[Inverted Double Pendulum]{\includegraphics[width = 0.33\textwidth, height=0.27\textwidth]{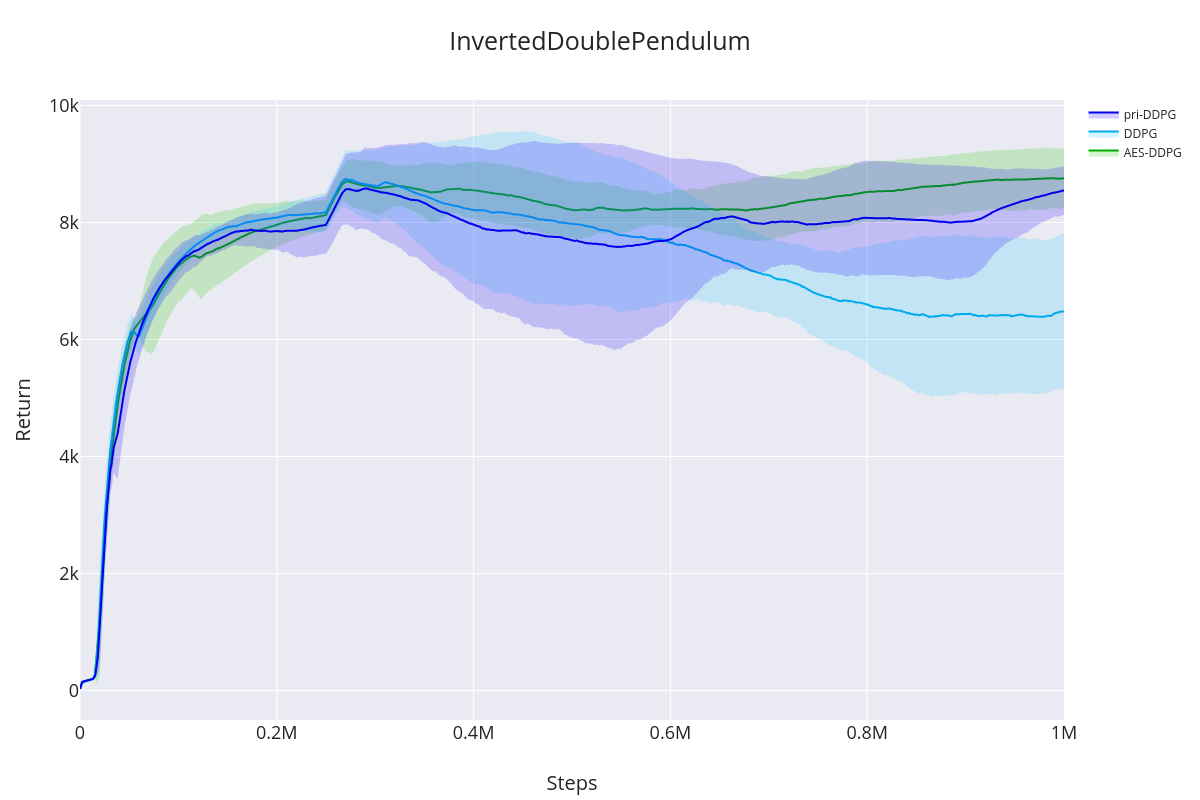}}
\subfigure[Reacher]{\includegraphics[width = 0.33\textwidth, height=0.27\textwidth]{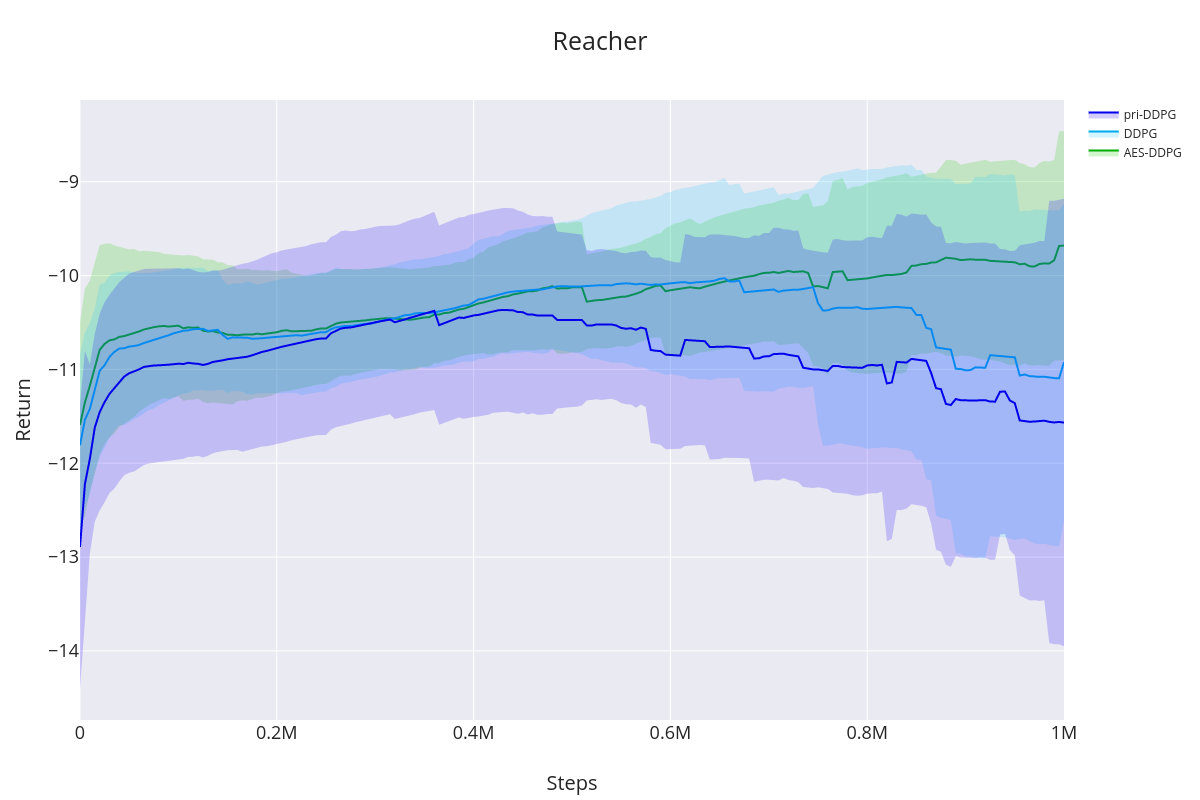}}\\
\subfigure[Hopper]{\includegraphics[width = 0.33\textwidth, height=0.27\textwidth]{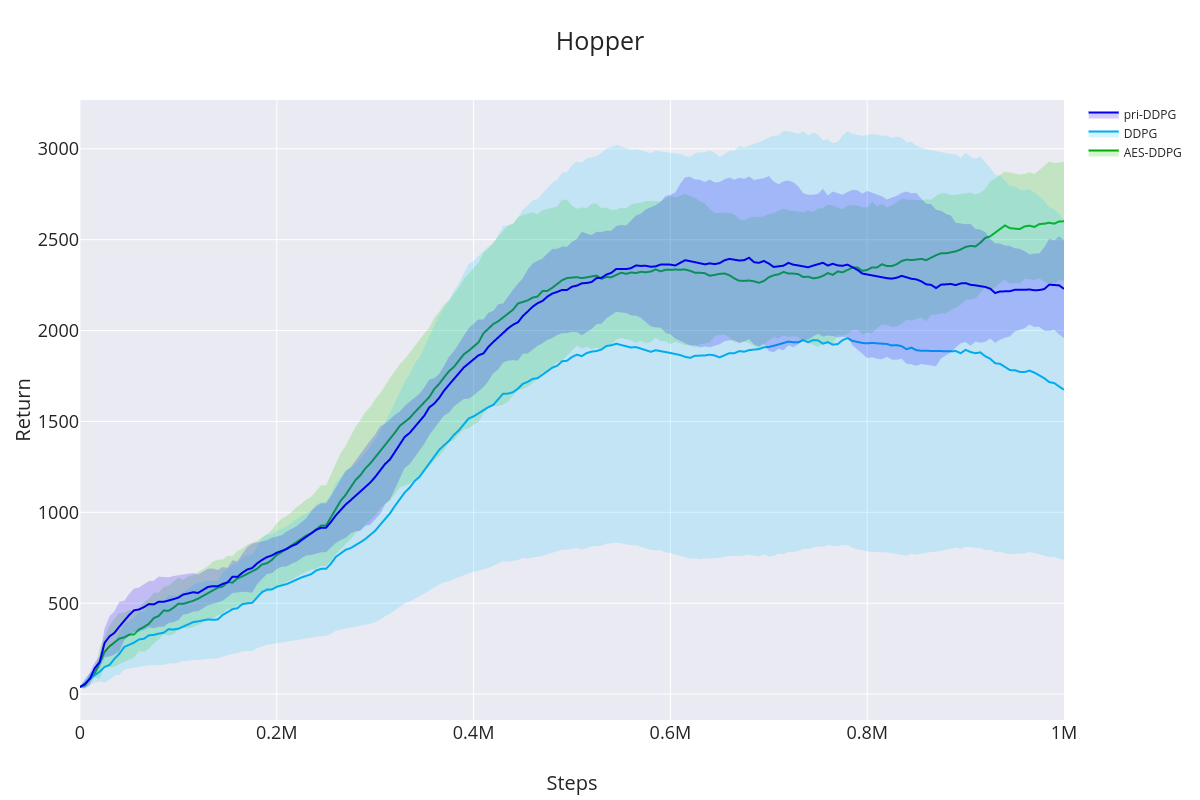}}
\subfigure[HalfCheetah]{\includegraphics[width = 0.33\textwidth, height=0.27\textwidth]{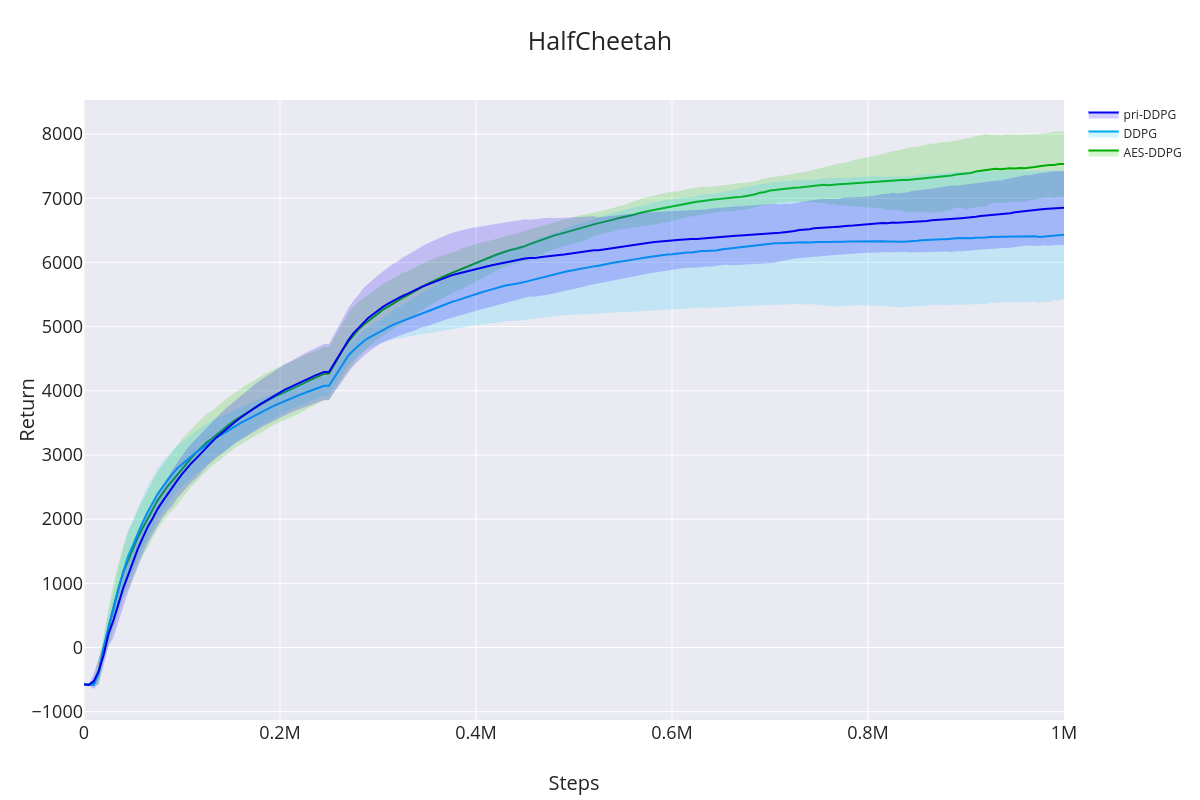}}
\subfigure[Walker2d]{\includegraphics[width = 0.33\textwidth, height=0.27\textwidth]{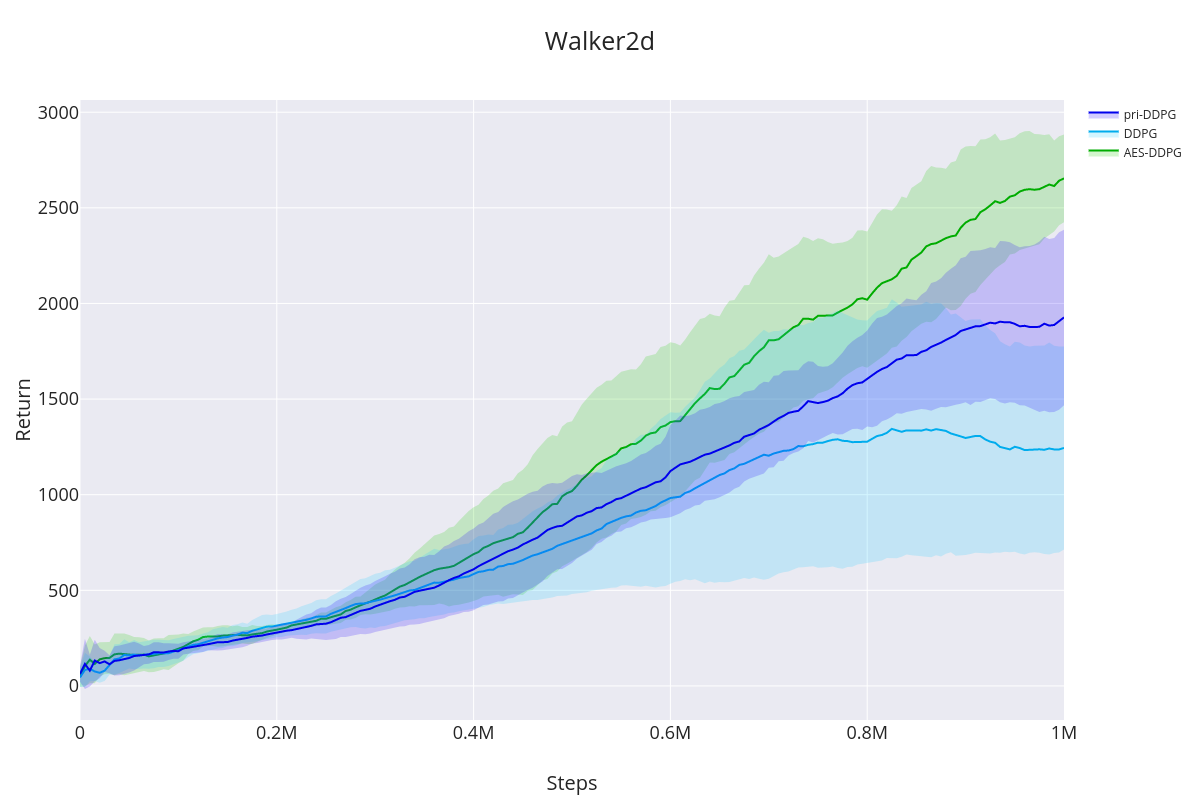}}\\
\subfigure[Ant]{\includegraphics[width = 0.33\textwidth, height=0.27\textwidth]{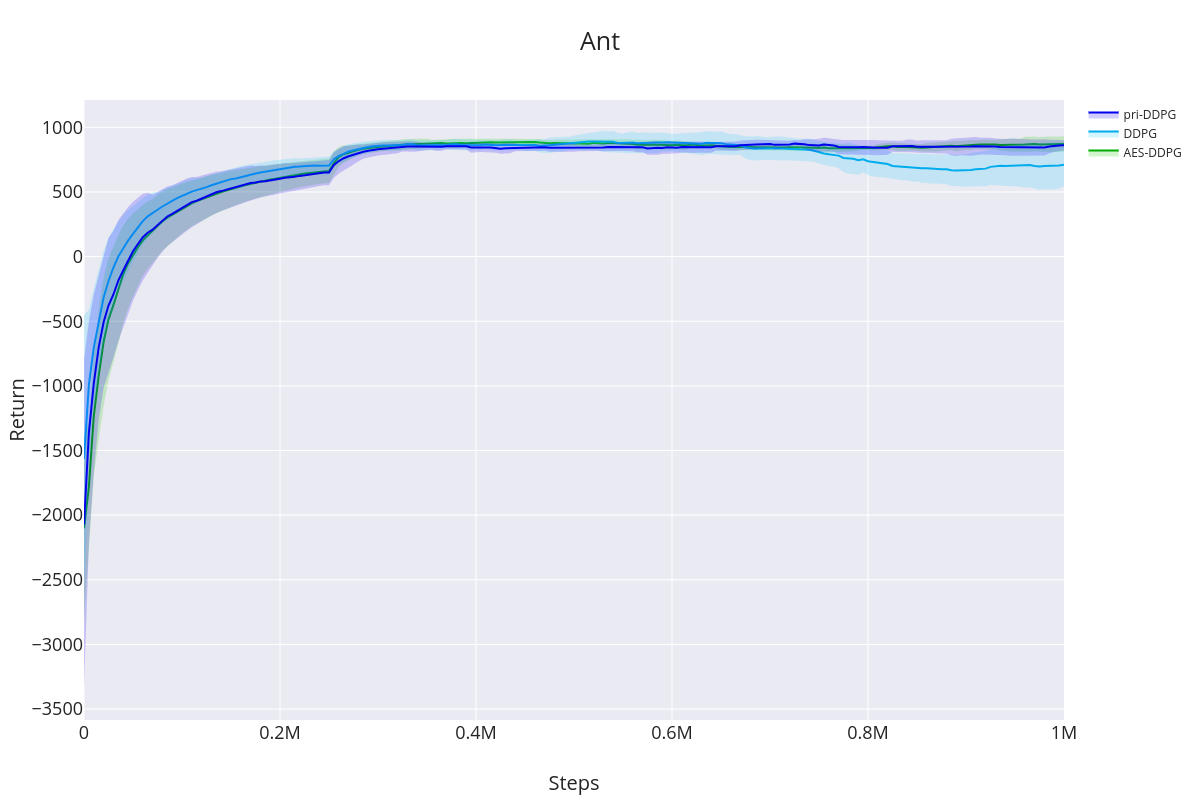}}
\subfigure[Humanoid]{\includegraphics[width = 0.33\textwidth, height=0.27\textwidth]{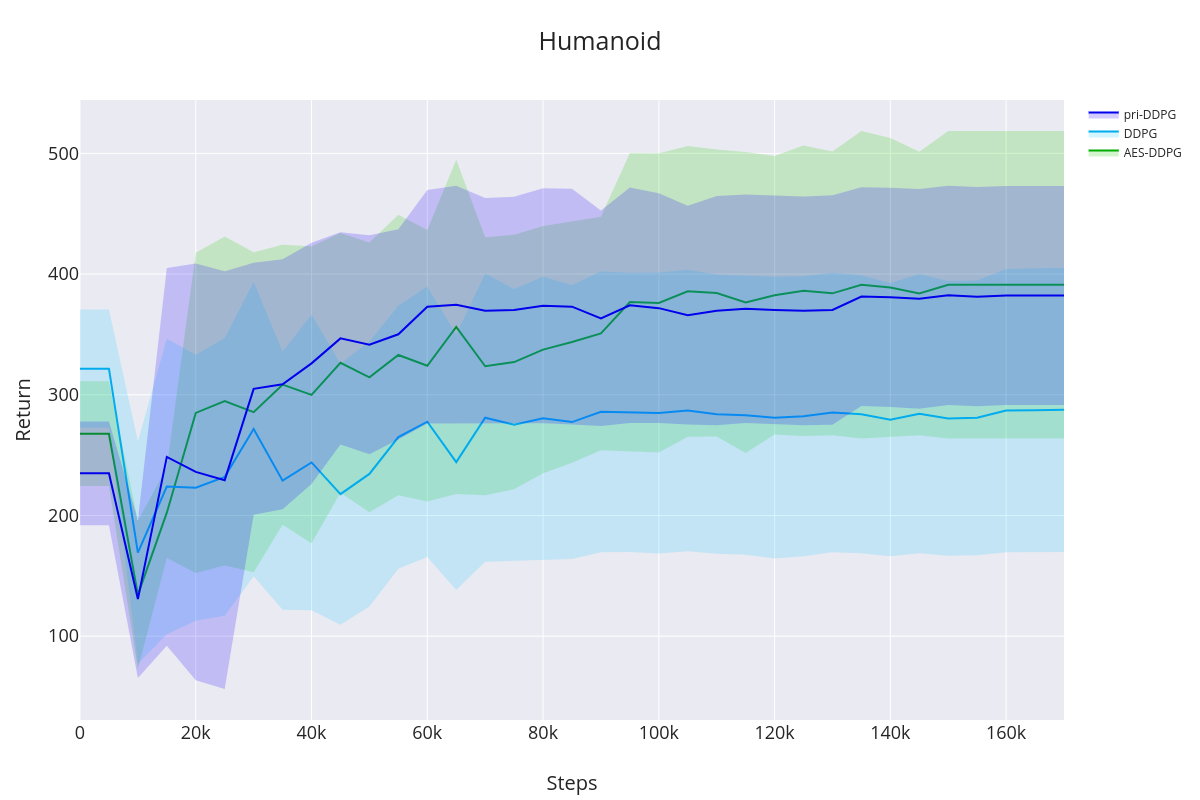}}
\label{fi4}
\caption{Performance of comparison methods with DDPG.}
\end{figure*} 

\begin{figure*}[ht]
\subfigure[Inverted Pendulum]{\includegraphics[width = 0.33\textwidth, height=0.27\textwidth]{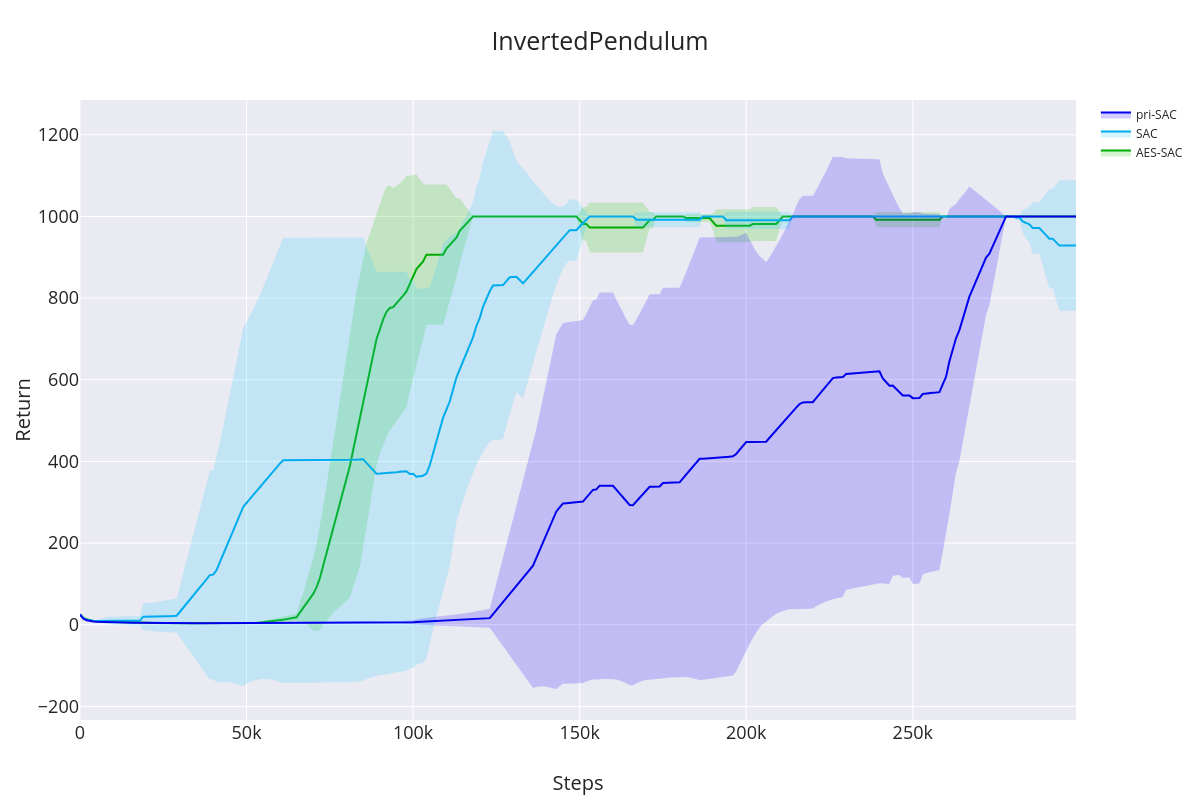}}
\subfigure[Inverted Double Pendulum]{\includegraphics[width = 0.33\textwidth, height=0.27\textwidth]{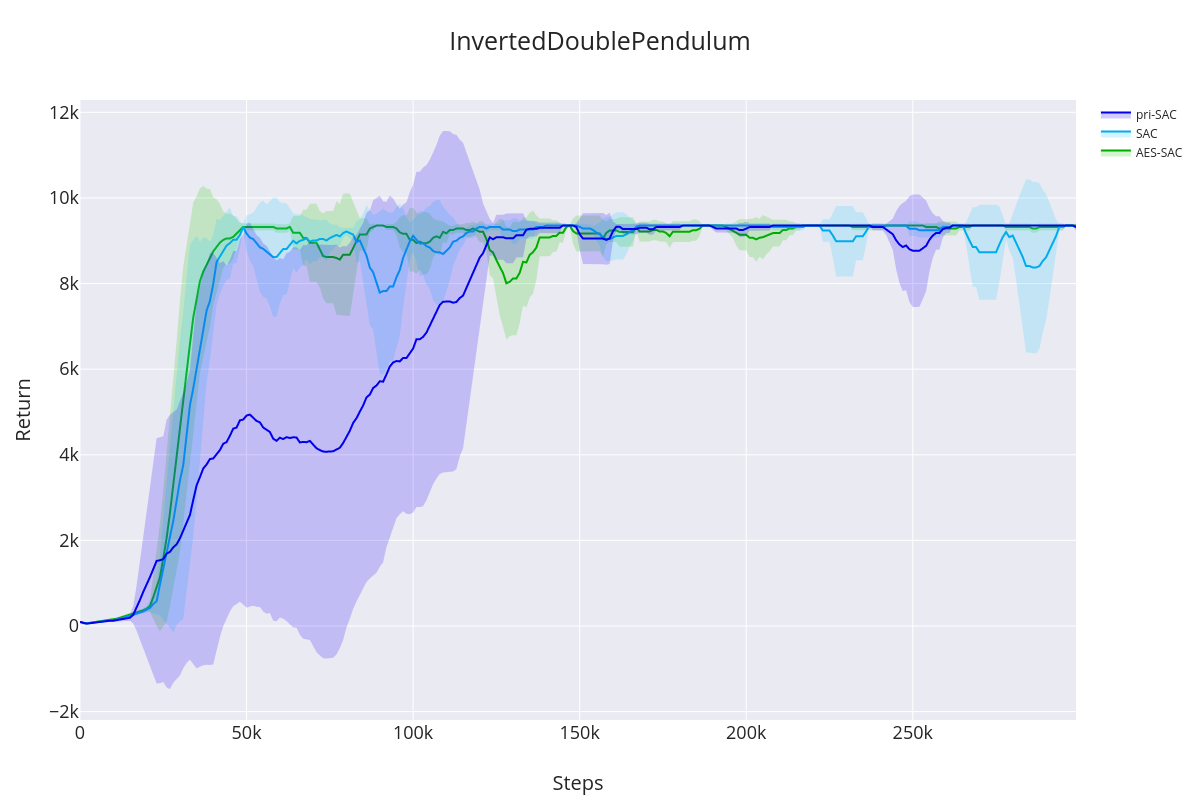}}
\subfigure[Reacher]{\includegraphics[width = 0.33\textwidth, height=0.27\textwidth]{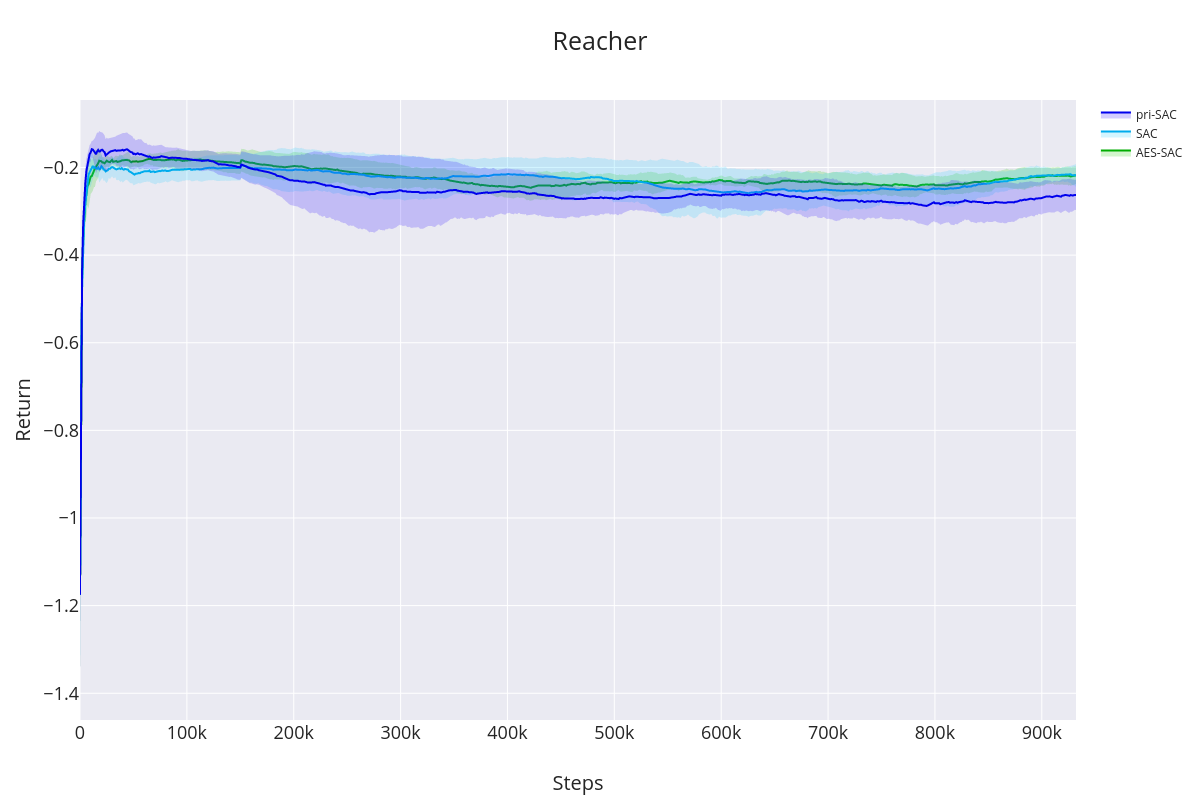}}\\
\subfigure[Hopper]{\includegraphics[width = 0.33\textwidth, height=0.27\textwidth]{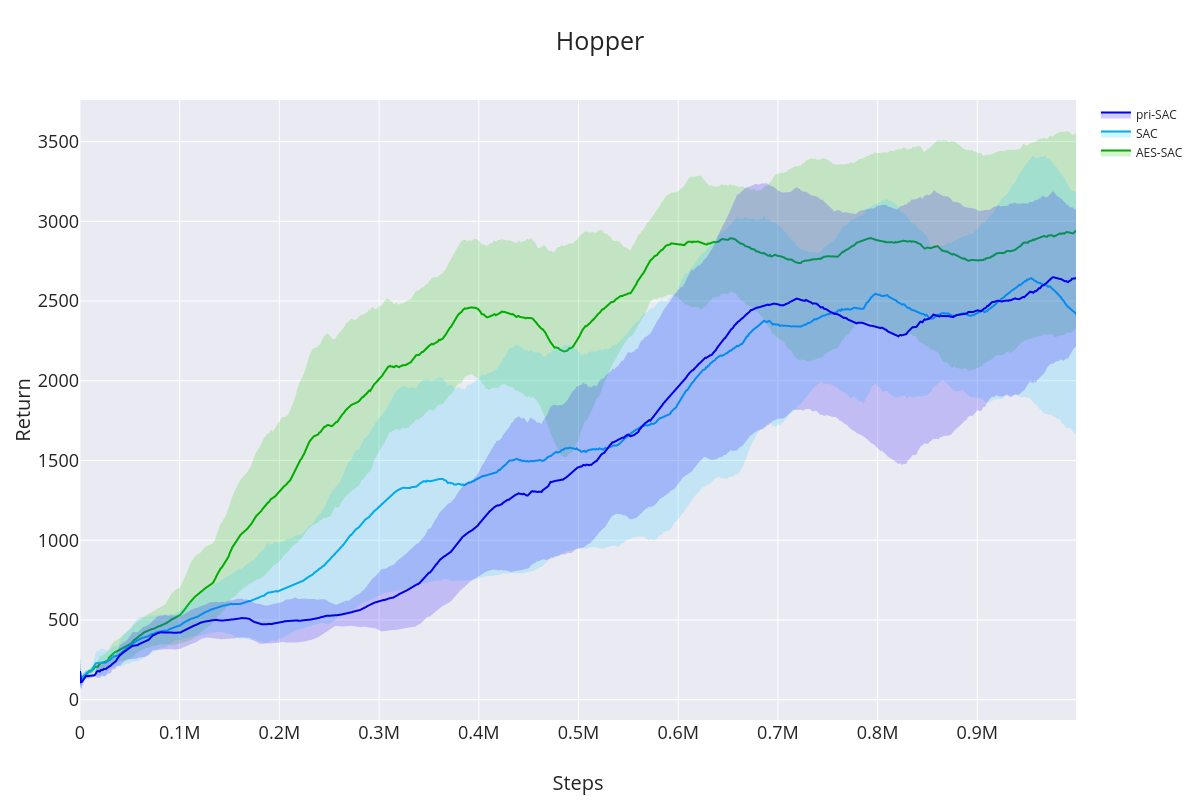}}
\subfigure[HalfCheetah]{\includegraphics[width = 0.33\textwidth, height=0.27\textwidth]{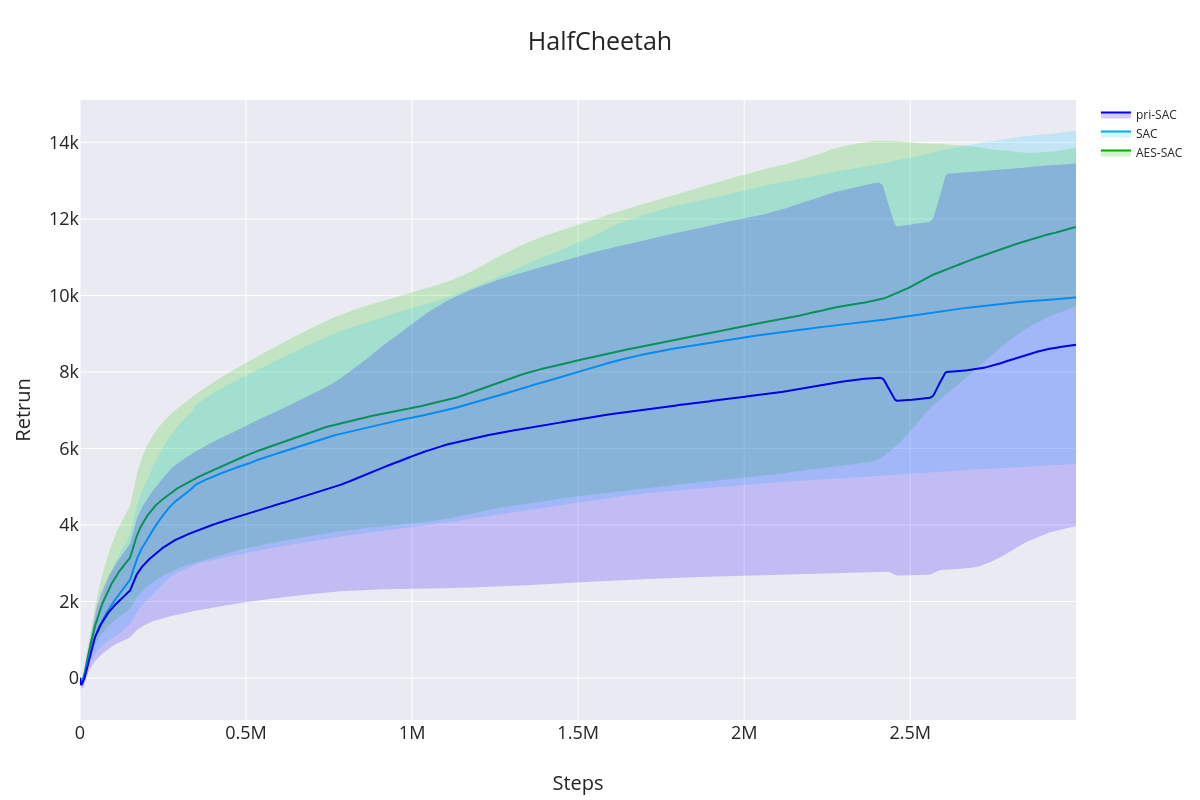}}
\subfigure[Walker2d]{\includegraphics[width = 0.33\textwidth, height=0.27\textwidth]{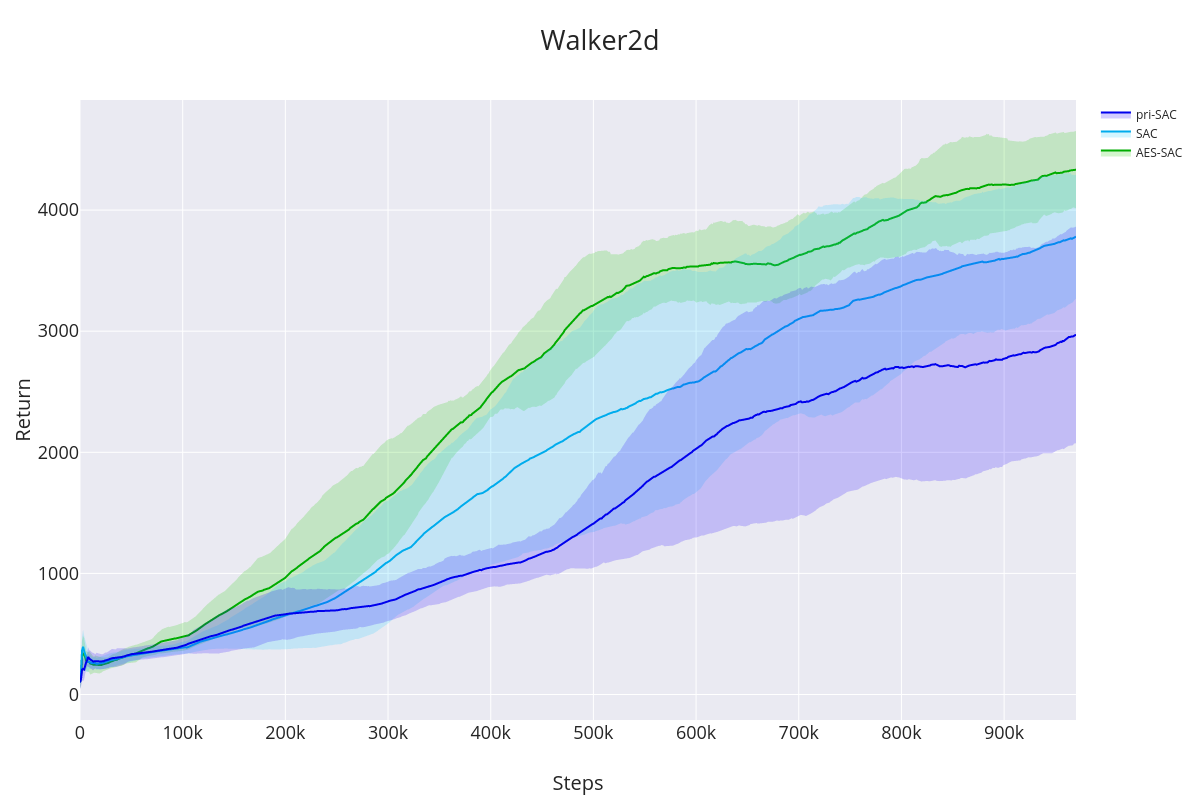}}\\
\subfigure[Ant]{\includegraphics[width = 0.33\textwidth, height=0.27\textwidth]{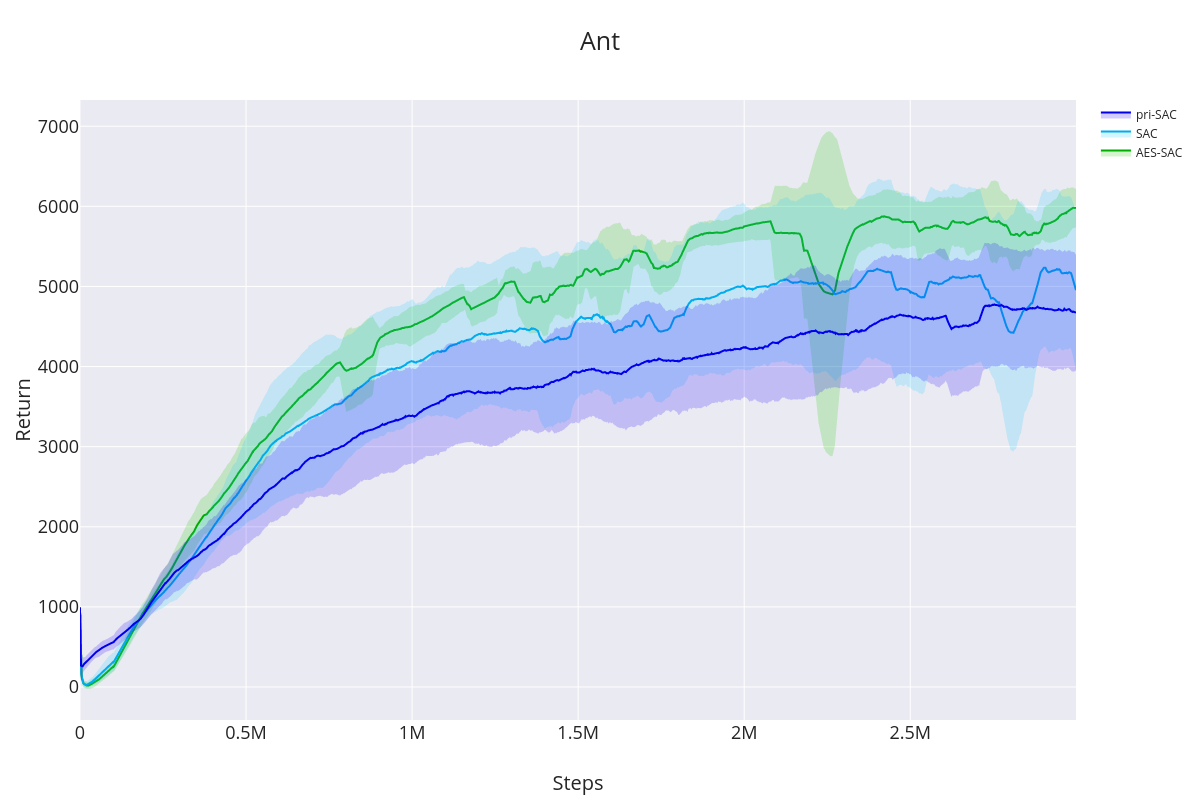}}
\subfigure[Humanoid]{\includegraphics[width = 0.33\textwidth, height=0.27\textwidth]{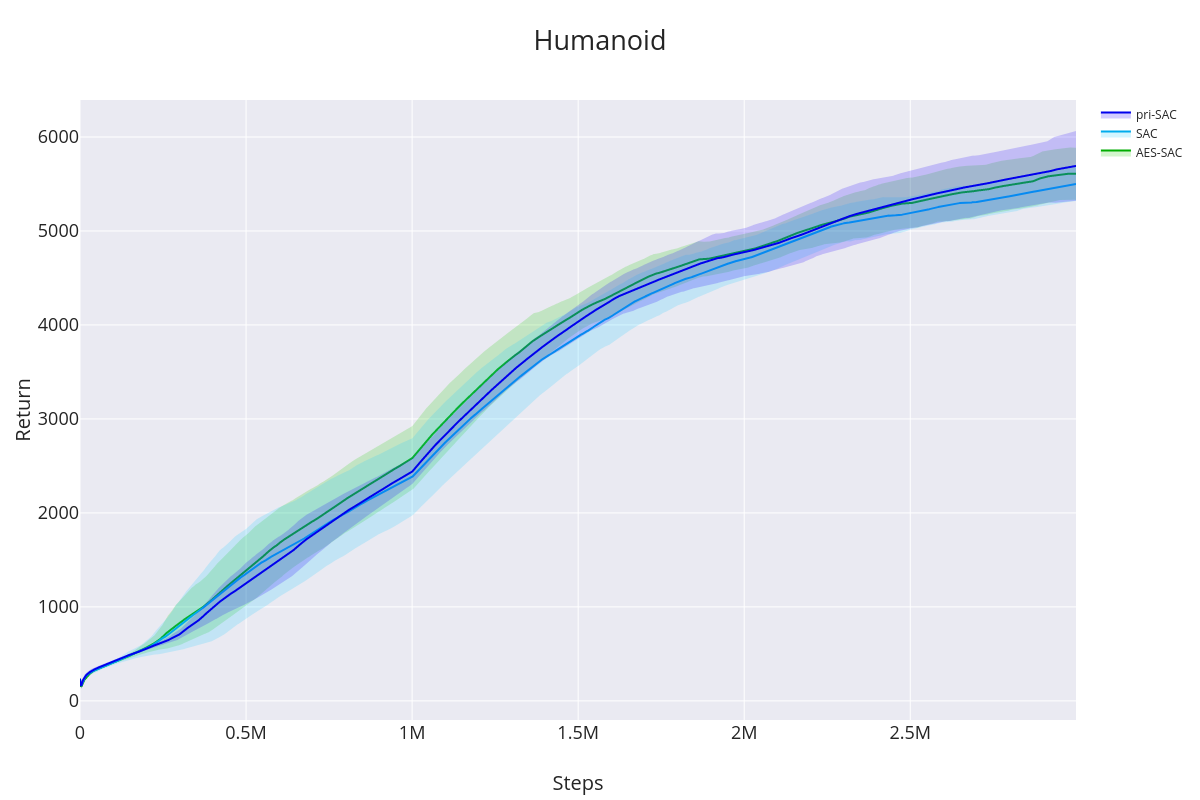}}
\label{fi5}
\caption{Performance of comparison methods with SAC.}
\end{figure*} 
Fig.~\ref{fi4} and Fig.~\ref{fi5} report the mean episodic return. From Fig.~\ref{fi4}, we can see that pri-DDPG performs generally better than DDPG, and AES-DDPG achieves improved performance compared to pri-DDPG. Specifically, for relatively simpler environments (Inverted Pendulum, Inverted Double Pendulum, Reacher, Hopper), AES-DDPG performs slightly better than pri-DDPG. For HalfCheetah and Walker2d, which are more complex tasks, AES-DDPG achieves a larger improvement compared to pri-DDPG. For Ant and Humanoid, which are the most complex ones, DDPG is unable to learn a good policy. As a result, the performance of AES-DDPG is unstable. AES-DDPG performs similarly to pri-DDPG on Ant, and slightly better than pri-DDPG on Humanoid. The improvement achieved by AES is not limited to sample efficiency, but involves also learning stability, robustness and final performance. AES-DDPG achieves the highest final episodic return on all the environments with slight fluctuations across the learning steps and little variation over the different seeds.   

Similar trends can be observed with SAC as shown in Fig.~\ref{fi5}. It can be seen that SAC-AES performs clearly better than SAC and pri-SAC on more complex environments (Hopper, HalfCheetah, Walker2d, Ant), while SAC-AES performs comparablly to SAC on simpler environments (Inverted Pendulum, Inverted Double Pendulum, Reacher). A possible reason here is that SAC alone is sufficient to achieve near-optimal performance on these environments due to their simplicity, and there is litte room for improvement. On the other hand, we observe that pri-SAC based on temporal difference error does not perform better than SAC with general uniform sampling. A potential problem could be that Prio depends on the objective function being optimised as SAC, unlike DDPG, modifies the policy gradient objective function by adding an additional term of the policy entropy. On the contrary, AES considers the objective function as a whole regardless of its components. Note that, SAC-AES outperforms SAC on the complex environment Humanoid, but its performance is comparable to pri-SAC. An explanation for this result can be related to~\cite{de2018experience}, who empirically demonstrated that the performance of heuristic experience selection method such as Prio depends on the characteristics of the control problem at hand. In addition to this assessment, we have also observed that the performance of heuristic experience selection methods also depends on the objective function defined by the underlhying RL algorithm. These results clearly demonstrate the ability of AES to adapt to different environments using different RL algorithms. 
\section{Conclusion and Discussion}\label{sec7}
In this paper, we have proposed an adaptive variance reduction methodology for policy gradient learning through experience replay, which has been framed as an online optimisation problem. AES is generic and does not modify the gradient estimate making it possible to integrate different existing methods to reduce the variance further. We demonstrate this claim by employing AES in two existing RL algorithms i.e., SAC and DDPG, that can be considered as gradient based VR policy gradient methods. We have empirically shown that AES improves the learning performance of SAC and DDPG compared to standard and prioritised experience replay. We also also provided theoretical analysis and justification for guaranteeing variance reduction within our framework. 

In future work, different approaches for handling the non-stationarity caused by RL experiences overwriting could be further explored. An alternative approach would use online unsupervised clustering algorithm to estimate the experience replay distribution directly in the experience space rather than imposing sampling distribution over their index. Overwriting would then update the density estimation (clustering) model of the experiences. This will prevents abrupt change in the sampling distribution and standard online learning can then be applied to find the best sampling distribution reducing the variance the most. On the application aspect, AES could be explored with multi-agent RL where the variance is known to be high and alleviating it is essential.  
\bibliographystyle{IEEEtran}
\bibliography{bibio}
\newpage
\onecolumn
\appendices
\label{appendix:a_proofs}
\section{Proofs}
\subsection{Proof of Lemma 1}
\label{appendix:a_proofs_1}
For simplicity, we write ${\bm\theta}_t$ as ${\bm\theta}$. Note that the output of $\omega_i^{\bm\theta}$ and $R$ are scalars. Therefore, we have the following equation:
\begin{equation}\label{eq:proof_lemma1}
d_t\left(i\right)={\left|{\omega_i^{\bm\theta}}{g_i^{\bm\theta}}\right|}^2={\left\|\omega\left(\tau_i|\pi_{\bm\theta}, \pi_{{\bm\theta}_{l\left(i\right)}}\right)\nabla \log p(\tau_i|\pi_{\bm\theta})R(\tau_i)\right\|}^2 = {\omega^2\left(\tau_i|\pi_{\bm\theta}, \pi_{{\bm\theta}_{l\left(i\right)}}\right)} {R^2(\tau_i)} {\left\|\nabla \log p(\tau_i|\pi_{\bm\theta})\right\|}^2 
\end{equation}
According to Eq.~\ref{eq:dist_tau} and Assumption 1, we have
\begin{equation}\label{eq:proof_lemma1_1}
\omega\left(\tau_i|\pi_{\bm\theta}, \pi_{{\bm\theta}_{l\left(i\right)}}\right) = \prod_{t=0}^{H-1}\frac{\pi_{\bm\theta}\left({\bm a}_{t}|{\bm s}_{t}\right)}{\pi_{{\bm\theta}_{l\left(i\right)}}\left({\bm a}_{t}|{\bm s}_{t}\right)} \leq \prod_{t=0}^{H-1} {\frac{1}{\beta}} = \frac{1}{\beta^H}
\end{equation}
Eq.~\ref{eq:dist_tau} and Assumption 2 lead to
\begin{equation}\label{eq:proof_lemma1_2}
{\left\|\nabla \log p(\tau_i|\pi_{\bm\theta})\right\|} = {\left\|\sum_{t=0}^{H-1}{\nabla\log \pi_{\bm\theta}({\bm a}_t|{\bm s}_t)}\right\|} \leq \sum_{t=0}^{H-1}{\left\|{\nabla\log \pi_{\bm\theta}({\bm a}_t|{\bm s}_t)}\right\|} \leq HL
\end{equation}
Using Assumption 3 we can obtain
\begin{equation}\label{eq:proof_lemma1_3}
R\left(\tau_i\right)=\sum_{t=0}^{H-1}\gamma^tr({\bm s}_t, {\bm a}_t) \leq \zeta \sum_{t=0}^{H-1}\gamma^t = \frac{1-\gamma^{H}}{1-\gamma}\zeta
\end{equation}
By substituting Inequalities (\ref{eq:proof_lemma1_1}), (\ref{eq:proof_lemma1_2}) and (\ref{eq:proof_lemma1_3}) to Eq.~\ref{eq:proof_lemma1}, we obtain
\begin{equation}\label{eq:proof_lemma1_final}
d_t\left(i\right) \leq {\left[\frac{\zeta(1-\gamma^{H})}{\beta^{H}(1-\gamma)}HL\right]}^2
\end{equation}
Thus, the proof of Lemma 1.
\subsection{Proof of Corollary 3}
\label{appendix:a_proofs_corollary_3}
Using Corollary 2, we have
\begin{equation}\label{eq:proof_corollary_3_1}
\E{\frac{1}{{|\mathcal{B}|}^2}\left(\sum_{t=1}^{m}{f_t\left(\tilde{\bm p}_t\right)}-\min_{{\bm p}\in\Delta}{\sum_{t=1}^{m}{f_t\left({\bm p}\right)}}\right)} \leq 74{\left[\frac{\zeta(1-\gamma^{H})}{\beta^{H}(1-\gamma)}HL\right]}^2 {|\mathcal{B}|}^{\frac{1}{3}}{m}^{\frac{2}{3}}
\end{equation}
It follows that 
\begin{equation}\label{eq:proof_corollary_3_2}
\sum_{k=1}^{E}{\frac{1}{m}\E{\frac{1}{{|\mathcal{B}|}^2}\left(\sum_{t=1}^{m}{f_t\left(\tilde{\bm p}_t\right)}-\min_{{\bm p}\in\Delta}{\sum_{t=1}^{m}{f_t\left({\bm p}\right)}}\right)}} \leq 74{\left[\frac{\zeta(1-\gamma^{H})}{\beta^{H}(1-\gamma)}HL\right]}^2 {|\mathcal{B}|}^{\frac{1}{3}}{m}^{-\frac{1}{3}}E
\end{equation}
Setting $m=E/C^2$ leads to sub-linear regret bound with respect to $E$:

\begin{equation}\label{skad}
{\sum_{k=1}^{E}{\frac{1}{m}\E{\frac{1}{{|\mathcal{B}|}^2}\left(\sum_{t=1}^{m}{f_t\left(\tilde{\bm p}_t\right)}-\min_{{\bm p}\in\Delta}{\sum_{t=1}^{m}{f_t\left({\bm p}\right)}}\right)}}}
\leq\mathcal{O}(|\mathcal{B}|^{1/3}C^{2/3}E^{2/3})
\end{equation}

Under assumption 4 and with $m=E/C^2$, we have $T<C^2\frac{|\mathcal{B}|}{|\Psi|}$. The regret bound can be written as follows:

\begin{equation}
\frac{1}{E}{\sum_{k=1}^{E}{\frac{1}{m}\E{\frac{1}{{|\mathcal{B}|}^2}\left(\sum_{t=1}^{m}{f_t\left(\tilde{\bm p}_t\right)}-\min_{{\bm p}\in\Delta}{\sum_{t=1}^{m}{f_t\left({\bm p}\right)}}\right)}}}
\leq\mathcal{O}(\frac{|\mathcal{B}|^{1/3}C^{2/3}}{E^{1/3}})
\end{equation}
Thus, Corollary 3 is proven.
\subsection{Proof of Theorem 1}
Recall that for AES we reset ${\bm p}$ every $M$ policy update steps. Thus, we divide the policy update steps into sequential sets each with $M$ steps, and use ${\mathcal{T}}_m$ to denote the $m^{\mathrm{th}}$ set where $m \in [1, 2, ... , \floor{T/M}+1]$. It follows up that:
\begin{align}\label{proosf1}
   \nonumber \frac{1}{{|\mathcal{B}|}^2}\E{\left(\sum_{t=1}^{T}{f_t\left(\tilde{\bm p}_t\right)}-{\sum_{t=1}^{T}{\min_{{\bm p} \in \Delta}{f_t\left({\bm p}\right)}}}\right)}=&\frac{1}{{|\mathcal{B}|}^2}\sum_{m=1}^{\floor{\frac{T}{M}}+1}{\E{\bigg(\sum_{t\in \mathcal{T}_m}f_t(\tilde{\bm p}_t)-\min_{{\bm p}\in \Delta}\sum_{t\in \mathcal{T}_m}f_t({\bm p})\bigg)}}\\
   +\frac{1}{{|\mathcal{B}|}^2}&\sum_{m=1}^{\floor{\frac{T}{M}}+1}{\E{\bigg(\min_{{\bm p}\in \Delta}\sum_{t\in \mathcal{T}_m}f_t({\bm p})-\sum_{t\in \mathcal{T}_m}\min_{{\bm p}\in \Delta}f_t({\bm p})\bigg)}}
\end{align}
Using Lemma 2, we can bound the second term of Eq.~\eqref{proosf1}:

\begin{equation}\label{va4}
\frac{1}{{|\mathcal{B}|}^2}\sum_{m=1}^{\floor{\frac{T}{M}}+1}E\bigg[\min_{{\bm p}\in \Delta}\sum_{t\in \mathcal{T}_m}f_t({\bm p})-\sum_{t\in \mathcal{T}_m}\min_{{\bm p}\in \Delta}f_t({\bm p})\bigg]\leq{\mathcal{O}(\frac{T}{\sqrt{|\mathcal{B}|M}})}
\end{equation}

Following Theorem 7 in~\cite{borsos2018online}, we can bound first part of Eq.~\eqref{proosf1} as follows:
\begin{align}\label{proosf1b}
   \nonumber {\frac{1}{{|\mathcal{B}|}^2}\E{\bigg(\sum_{t\in \mathcal{T}_m}f_t(\tilde{\bm p}_t)-\min_{{\bm p}\in \Delta}\sum_{t\in \mathcal{T}_m}f_t({\bm p})\bigg)}} \leq \kappa\sum_{t=1}^M \sum_{i=1}^{|\mathcal{B}|}||\omega_i^{\bm\theta_{t}}g_i^{\bm\theta_{t}} ||^2+27\frac{\left[\frac{\zeta(1-\gamma^{H})}{\beta^{H}(1-\gamma)}HL\right]}{\sqrt{\kappa|\mathcal{B}|}}\sum_{t=1}^M\sum_{i=1}^{|\mathcal{B}|}||\omega_i^{\bm\theta_{t}}g_i^{\bm\theta_{t}} ||+\\\frac{44|\mathcal{B}|\left[\frac{\zeta(1-\gamma^{H})}{\beta^{H}(1-\gamma)}HL\right]^2}{\kappa}  
\end{align}
where $|\mathcal{T}_m|=M$ $\forall m$. Using the same steps used in lemma 2 to move from Eq.~\eqref{vavarsg2} to Eq.~\eqref{varsgnew4}, we can bound Eq.~\eqref{proosf1b} as follows:

\begin{align}\label{proosf12}
   \nonumber {\frac{1}{{|\mathcal{B}|}^2}\E{\bigg(\sum_{t\in \mathcal{T}_m}f_t(\tilde{\bm p}_t)-\min_{{\bm p}\in \Delta}\sum_{t\in \mathcal{T}_m}f_t({\bm p})\bigg)}} \leq \frac{2K|\mathcal{B}|\kappa}{\beta^2}\sum_{t=1}^M \big(J({\bm\theta^{*}})-J({\bm\theta_{t}})\big)+\frac{27\sqrt{2K}\left[\frac{\zeta(1-\gamma^{H})}{\beta^{H}(1-\gamma)}HL\right]}{\beta\sqrt{\kappa}}\sum_{t=1}^M\sqrt{\big(J({\bm\theta^{*}})-J({\bm\theta_{t}})\big)}+\\\frac{44|\mathcal{B}|\left[\frac{\zeta(1-\gamma^{H})}{\beta^{H}(1-\gamma)}HL\right]^2}{\kappa}  
\end{align}

Using the steps used in lemma 2 to move from Eq.~\eqref{varsgnew4} to Eq.~\eqref{vavarsg3}, we can bound Eq.~\eqref{proosf1b} as follows:

\begin{align}\label{proosf13}
   \nonumber {\frac{1}{{|\mathcal{B}|}^2}\E{\bigg(\sum_{t\in \mathcal{T}_m}f_t(\tilde{\bm p}_t)-\min_{{\bm p}\in \Delta}\sum_{t\in \mathcal{T}_m}f_t({\bm p})\bigg)}} \leq \frac{2K|\mathcal{B}|\kappa}{\beta^2}\sum_{t=1}^M \bigg(E\big[J({\bm\theta^{*}})-J({\bm\theta_{t}})\big]+2\frac{t(1-\gamma^{H})\zeta}{|\mathcal{B}|(1-\gamma)}\bigg)+\\\frac{27\sqrt{2K}\left[\frac{\zeta(1-\gamma^{H})}{\beta^{H}(1-\gamma)}HL\right]}{\beta\sqrt{\kappa}}\sum_{t=1}^M\sqrt{E\big[J({\bm\theta^{*}})-J({\bm\theta_{t}})\big]+2\frac{t(1-\gamma^{H})\zeta}{|\mathcal{B}|(1-\gamma)}}+\frac{44|\mathcal{B}|\left[\frac{\zeta(1-\gamma^{H})}{\beta^{H}(1-\gamma)}HL\right]^2}{\kappa}  
\end{align}

For these steps, Assumption 3 and 5  need to hold. Using Jensen's inequality and with a bit of algebra, we can obtain the following:
\begin{align}\label{proosf14}
   \nonumber {\frac{1}{{|\mathcal{B}|}^2}\E{\bigg(\sum_{t\in \mathcal{T}_m}f_t(\tilde{\bm p}_t)-\min_{{\bm p}\in \Delta}\sum_{t\in \mathcal{T}_m}f_t({\bm p})\bigg)}} \leq \frac{2K|\mathcal{B}|\kappa}{\beta^2}\bigg(\sum_{t=1}^M E\big[J({\bm\theta^{*}})-J({\bm\theta_{t}})\big]+2\frac{M(M+1)(1-\gamma^{H})\zeta}{|\mathcal{B}|(1-\gamma)}\bigg)+\\\frac{27\sqrt{2K}\left[\frac{\zeta(1-\gamma^{H})}{\beta^{H}(1-\gamma)}HL\right]}{\beta\sqrt{\kappa}}\bigg(\sum_{t=1}^M\sqrt{E\big[J({\bm\theta^{*}})-J({\bm\theta_{t}})\big]}+2\frac{(1-\gamma^{H})\zeta}{|\mathcal{B}|(1-\gamma)}(\frac{2M^{3/2}}{3}+\mathcal{O}(\sqrt{M}))\bigg)+\frac{44|\mathcal{B}|\left[\frac{\zeta(1-\gamma^{H})}{\beta^{H}(1-\gamma)}HL\right]^2}{\kappa}  
\end{align}

Under assumption 6 and given that the number of overwritten samples before resetting is less than the buffer size $M^2<|\mathcal{B}|$ (this condition is already needed for lemma 2), we use the steps of lemma 2 used to move from Eq.~\eqref{vavarsg3} to Eq.~\eqref{vavarsg31}:
\begin{align}\label{proosf15}
   \nonumber {\frac{1}{{|\mathcal{B}|}^2}\E{\bigg(\sum_{t\in \mathcal{T}_m}f_t(\tilde{\bm p}_t)-\min_{{\bm p}\in \Delta}\sum_{t\in \mathcal{T}_m}f_t({\bm p})\bigg)}} \leq \frac{2K|\mathcal{B}|\kappa}{\beta^2}\bigg(\sum_{t=1}^M (1-\xi\alpha)^{t-1}\big(J({\bm\theta^{*}})-J({\bm\theta_{0}})\big)+2\frac{(1-\gamma^{H})\zeta}{(1-\gamma)}\bigg)+\\\frac{27\sqrt{2K}\left[\frac{\zeta(1-\gamma^{H})}{\beta^{H}(1-\gamma)}HL\right]}{\beta\sqrt{\kappa}}\bigg(\sum_{t=1}^M\sqrt{(1-\xi\alpha)^{t-1}\big(J({\bm\theta^{*}})-J({\bm\theta_{0}})\big)}+2\frac{(1-\gamma^{H})\zeta}{(1-\gamma)}(\frac{2}{3\sqrt{M}}+\mathcal{O}(\frac{1}{M^{3/2}}))\bigg)+\frac{44|\mathcal{B}|\left[\frac{\zeta(1-\gamma^{H})}{\beta^{H}(1-\gamma)}HL\right]^2}{\kappa}  
\end{align}

Thus,
\begin{equation}\label{proosf16}
   {\frac{1}{{|\mathcal{B}|}^2}\E{\bigg(\sum_{t\in \mathcal{T}_m}f_t(\tilde{\bm p}_t)-\min_{{\bm p}\in \Delta}\sum_{t\in \mathcal{T}_m}f_t({\bm p})\bigg)}} \leq \mathcal{O}(\kappa |\mathcal{B}|)+\mathcal{O}(\frac{1}{\sqrt{\kappa M}})+\mathcal{O}(\frac{|\mathcal{B}|}{\kappa})
\end{equation}

Using Eq.~\eqref{proosf16} with Eq.~\eqref{va4}, we have:

\begin{equation}\label{Final1}
    \frac{1}{{|\mathcal{B}|}^2}\E{\left(\sum_{t=1}^{T}{f_t\left(\tilde{\bm p}_t\right)}-{\sum_{t=1}^{T}{\min_{{\bm p} \in \Delta}{f_t\left({\bm p}\right)}}}\right)}\leq \mathcal{O}(\frac{\kappa |\mathcal{B}|T}{M})+\mathcal{O}(\frac{T}{M\sqrt{\kappa M}})+\mathcal{O}(\frac{|\mathcal{B}|T}{\kappa M})+\mathcal{O}(\frac{T}{\sqrt{|\mathcal{B}|M}})
\end{equation}
Thus, we prove theorem 2.
\subsection{Proof of Lemma 2:}
Under assumption 1, 2, 3, 5 and 6, lemma 2 bounds regret between static and dynamic optimum within epoch:  
\begin{align}\label{proosf1c}
\nonumber &\min_{{\bm p}\in \Delta}\sum_{t\in \mathcal{T}_m}f_t({\bm p})-\sum_{t\in \mathcal{T}_m}\min_{{\bm p}\in \Delta}f_t({\bm p}) =\\&\min_p\sum_{t\in \mathcal{T}_m}\sum_{i=1}^{|\mathcal{B}|}\frac{||\omega_i^{\bm\theta_{t}}g_i^{\bm\theta_{t}} ||^2}{p(i)}-\sum_{t\in \mathcal{T}_m}\min_{p}\sum_{i=1}^{|\mathcal{B}|}\frac{||\omega_i^{\bm\theta_{t}}g_i^{\bm\theta_{t}} ||^2}{p(i)}
\end{align}
Taking $p^*=\argmin_{p\in \Delta}\sum_{t\in \mathcal{T}_m}\sum_{i=1}^{|\mathcal{B}|}\frac{||\omega_i^{\bm\theta_{t}}g_i^{\bm\theta_{t}} ||^2}{p(i)}$ and $p_t^*=\argmin_{p\in \Delta}\sum_{i=1}^{|\mathcal{B}|}\frac{||\omega_i^{\bm\theta_{t}}g_i^{\bm\theta_{t}} ||^2}{p(i)}$, Eq.~\eqref{proosf1c} can be expressed as:
\begin{align}\label{kls}
    \nonumber&\min_p\sum_{t\in \mathcal{T}_m}\sum_{i=1}^{|\mathcal{B}|}\frac{||\omega_i^{\bm\theta_{t}}g_i^{\bm\theta_{t}} ||^2}{p(i)}-\sum_{t\in \mathcal{T}_m}\min_{p}\sum_{i=1}^{|\mathcal{B}|}\frac{||\omega_i^{\bm\theta_{t}}g_i^{\bm\theta_{t}} ||^2}{p(i)}=\\&\sum_{t\in \mathcal{T}_m}\sum_{i=1}^{|\mathcal{B}|}||\omega_i^{\bm\theta_{t}}g_i^{\bm\theta_{t}} ||^2\bigg( \frac{1}{p^*(i)}-\frac{1}{p_t^*(i)}\bigg)
\end{align}
To find $p^*$, we solve the following optimisation problems:
\begin{align}\label{opt11}
  \nonumber &\min_{p\in \Delta}\sum_{t\in \mathcal{T}_m}\sum_{i=1}^{|\mathcal{B}|}\frac{||\omega_i^{\bm\theta_{t}}g_i^{\bm\theta_{t}} ||^2}{p(i)}\\ \nonumber &\sum_{i=1}^{E}p(i)=1
\\  &p(i)\geq 0, \quad i=1,...|\mathcal{B}|
\end{align}
By formulating the Lagrangian, setting its derivative to zero and using complementary slackness, we can show that:
\begin{equation}\label{solution1}
    p^*(i)=\frac{\sqrt{\sum_{t\in \mathcal{T}_m}||\omega_i^{\bm\theta_{t}}g_i^{\bm\theta_{t}} ||^2}}{\sum_{j=1}^{|\mathcal{B}|}\sqrt{\sum_{t\in \mathcal{T}_m}||\omega_j^{\bm\theta_{t}}g_j^{\bm\theta_{t}} ||^2}}
\end{equation}
Similarly, to find $p_t^*$, we solve the optimisation problem:
\begin{align}\label{opt12}
  \nonumber &\min_{p\in \Delta}\sum_{i=1}^{|\mathcal{B}|}\frac{||\omega_i^{\bm\theta_{t}}g_i^{\bm\theta_{t}} ||^2}{p(i)}\\ \nonumber &\sum_{i=1}^{E}p(i)=1
\\  &p(i)\geq 0, \quad i=1,...|\mathcal{B}|
\end{align}
By formulating the Lagrangian, setting its derivative to zero and using complementary slackness, we get:
\begin{equation}\label{solution2}
    p_t^*(i)=\frac{||\omega_i^{\bm\theta_{t}}g_i^{\bm\theta_{t}} ||}{\sum_{j=1}^{|\mathcal{B}|}||\omega_j^{\bm\theta_{t}}g_j^{\bm\theta_{t}} ||}
\end{equation}

By substituting $p^*$ and $p_t^*$ in Eq.~\eqref{kls} and doing a bit of algebra, we can express Eq.~\eqref{proosf1c} as :
\begin{equation}\label{varsg}
\min_{{\bm p}\in \Delta}\sum_{t\in \mathcal{T}_m}f_t({\bm p})-\sum_{t\in \mathcal{T}_m}\min_{{\bm p}\in \Delta}f_t({\bm p}) =\sum_{i=1}^{|\mathcal{B}|}\sum_{j=1}^{|\mathcal{B}|}\sqrt{\sum_{t\in \mathcal{T}_m}\sum_{t'\in \mathcal{T}_m}(||\omega_i^{\bm\theta_{t}}g_i^{\bm\theta_{t}} ||^2)(||\omega_j^{\bm\theta_{t'}}g_j^{\bm\theta_{t'}} ||^2)}-\sum_{t\in \mathcal{T}_m}\sum_{i=1}^{|\mathcal{B}|}\sum_{j=1}^{|\mathcal{B}|}(||\omega_i^{\bm\theta_{t}}g_i^{\bm\theta_{t}} ||)(||\omega_j^{\bm\theta_{t}}g_j^{\bm\theta_{t}} ||)
\end{equation}
Using Lemma 1, we can get bound Eq.~\eqref{varsg}: 

\begin{equation}\label{vavarsg2}
\min_{{\bm p}\in \Delta}\sum_{t\in \mathcal{T}_m}f_t({\bm p})-\sum_{t\in \mathcal{T}_m}\min_{{\bm p}\in \Delta}f_t({\bm p}) \leq|\mathcal{B}|{\left[\frac{\zeta(1-\gamma^{H})}{\beta^{H}(1-\gamma)}HL\right]}\sum_{i=1}^{|\mathcal{B}|}\sqrt{|\mathcal{T}_m|\sum_{t\in \mathcal{T}_m}||\omega_i^{\bm\theta_{t}}g_i^{\bm\theta_{t}} ||^2}
\end{equation}

Assuming that no over-witting is occurring before resetting $\mathcal{T}_m$, we can consider the RL as a SGD-based optimisation problem for the data in the replay buffer, hence, $J(\bm\theta)=\frac{1}{|\mathcal{B}|}\sum_{j=1}^{|\mathcal{B}|}J_i(\bm\theta)$. Assume that the variance of the gradient is zeros for optimum policy $E[||g_j^{\bm\theta^*} ||^2]=0$, where $\bm\theta^*=argmax_{\bm\theta}J(\bm\theta)$· Under assumption 5, we can use lemma 1 in~\cite{lei2019stochastic} to show that:

\begin{equation}\label{varsgnew2}
\min_{{\bm p}\in \Delta}\sum_{t\in \mathcal{T}_m}f_t({\bm p})-\sum_{t\in \mathcal{T}_m}\min_{{\bm p}\in \Delta}f_t({\bm p}) \leq\frac{|\mathcal{B}|\sqrt{2K}}{\beta}{\left[\frac{\zeta(1-\gamma^{H})}{\beta^{H}(1-\gamma)}HL\right]}\sum_{i=1}^{|\mathcal{B}|}\sqrt{|\mathcal{T}_m|\sum_{t\in \mathcal{T}_m}\big(J_i(\bm\theta^{*})-J_i(\bm\theta_{t})\big)}
\end{equation}

Using Jensen's inequality:

\begin{equation}\label{varsgnew4}
\min_{{\bm p}\in \Delta}\sum_{t\in \mathcal{T}_m}f_t({\bm p})-\sum_{t\in \mathcal{T}_m}\min_{{\bm p}\in \Delta}f_t({\bm p}) \leq\frac{|\mathcal{B}|\sqrt{2K|\mathcal{B}|}}{\beta}{\left[\frac{\zeta(1-\gamma^{H})}{\beta^{H}(1-\gamma)}HL\right]}\sqrt{|\mathcal{T}_m|\sum_{t\in \mathcal{T}_m}\big(J({\bm\theta^{*}})-J({\bm\theta_{t}})\big)}
\end{equation}

Taking into account the overwriting occurring within the $|\mathcal{T}_m|$ steps, Eq.~\eqref{varsgnew4} becomes:
\begin{equation}\label{varsgnew4dgf}
\min_{{\bm p}\in \Delta}\sum_{t\in \mathcal{T}_m}f_t({\bm p})-\sum_{t\in \mathcal{T}_m}\min_{{\bm p}\in \Delta}f_t({\bm p}) \leq\frac{|\mathcal{B}|\sqrt{2K|\mathcal{B}|}}{\beta}{\left[\frac{\zeta(1-\gamma^{H})}{\beta^{H}(1-\gamma)}HL\right]}\sqrt{|\mathcal{T}_m|\sum_{t\in \mathcal{T}_m}\big(\tilde{J}({\bm\theta^{*}})-\tilde{J}({\bm\theta_{t}})\big)}
\end{equation}

where $\tilde{J}(\bm\theta_t)$ denotes the objective function with overwriting. We can bound the new objective $\tilde{J}(\bm\theta_t)$ as follows:
\begin{equation}
 J(\bm\theta_t)- t\frac{\max_\tau R(\tau)-\min_\tau R(\tau) }{|\mathcal{B}|} \leq\tilde{J}(\bm\theta_t)\leq J(\bm\theta_t)+t\frac{\max_\tau R(\tau)-\min_\tau R(\tau) }{|\mathcal{B}|}
\end{equation}

Using Assumption 3, we have:  

\begin{equation}\label{something}
 J(\bm\theta_t)-\frac{t(1-\gamma^{H})\zeta}{|\mathcal{B}|(1-\gamma)} \leq\tilde{J}(\bm\theta_t)\leq J(\bm\theta_t)+\frac{t(1-\gamma^{H})\zeta}{|\mathcal{B}|(1-\gamma)}
\end{equation}

Setting $|\mathcal{T}_m|=M$ $\forall m$ and using Eq.~\eqref{something}, we can bound Eq.~\eqref{varsgnew4} as follows:
\begin{equation}\label{varsgnew42}
\min_{{\bm p}\in \Delta}\sum_{t\in \mathcal{T}_m}f_t({\bm p})-\sum_{t\in \mathcal{T}_m}\min_{{\bm p}\in \Delta}f_t({\bm p}) \leq\frac{|\mathcal{B}|\sqrt{2K|\mathcal{B}|}}{\beta}{\left[\frac{\zeta(1-\gamma^{H})}{\beta^{H}(1-\gamma)}HL\right]}\sqrt{M\sum_{t=1}^M\bigg(J(\bm\theta^{*})-J(\bm\theta_{t})+2\frac{t(1-\gamma^{H})\zeta}{|\mathcal{B}|(1-\gamma)}\bigg)}
\end{equation}

Taking expectation over $\{\bm\theta_{t}\}_{t=1}^M$ and using Jensen's inequality, we have
\begin{equation}\label{varsgnew5}
E\bigg[\min_{{\bm p}\in \Delta}\sum_{t\in \mathcal{T}_m}f_t({\bm p})-\sum_{t\in \mathcal{T}_m}\min_{{\bm p}\in \Delta}f_t({\bm p})\bigg] \leq\frac{|\mathcal{B}|\sqrt{2K|\mathcal{B}|}}{\beta}{\left[\frac{\zeta(1-\gamma^{H})}{\beta^{H}(1-\gamma)}HL\right]}\sqrt{M\bigg(\sum_{t=1}^M E\big[J({\bm\theta^{*}})-J({\bm\theta_{t}})\big]+\frac{M(M+1)(1-\gamma^{H})\zeta}{|\mathcal{B}|(1-\gamma)}\bigg)}
\end{equation}

Under assumption 6, we can use Theorem 4 in~\cite{lei2019stochastic} to show that:
\begin{equation}\label{vavarsg3}
E\bigg[\min_{{\bm p}\in \Delta}\sum_{t\in \mathcal{T}_m}f_t({\bm p})-\sum_{t\in \mathcal{T}_m}\min_{{\bm p}\in \Delta}f_t({\bm p})\bigg] \leq\frac{|\mathcal{B}|\sqrt{2K|\mathcal{B}|}}{\beta}{\left[\frac{\zeta(1-\gamma^{H})}{\beta^{H}(1-\gamma)}HL\right]}\sqrt{M\sum_{t=1}^M (1-\xi\alpha)^{t-1}\big(J({\bm\theta^{*}})-J({\bm\theta_{0}})\big)+\frac{M^2(M+1)(1-\gamma^{H})\zeta}{|\mathcal{B}|(1-\gamma)}}
\end{equation}
where $\alpha_t=\alpha\leq \xi/K^2$. Assume $0\leq(1-\xi\alpha)<1$, hence $\alpha<1/ \xi$ which implies a condition on the learning rate according to the smoothness degree of the objective function $\alpha<1/K$. Given that number of overwritten samples before resetting is much less than the buffer size $M^2<|\mathcal{B}|$:

\begin{equation}\label{vavarsg31}
E\bigg[\min_{{\bm p}\in \Delta}\sum_{t\in \mathcal{T}_m}f_t({\bm p})-\sum_{t\in \mathcal{T}_m}\min_{{\bm p}\in \Delta}f_t({\bm p})\bigg]\leq\frac{|\mathcal{B}|\sqrt{2K|\mathcal{B}|}}{\beta}{\left[\frac{\zeta(1-\gamma^{H})}{\beta^{H}(1-\gamma)}HL\right]}\sqrt{M\sum_{t=1}^M (1-\xi\alpha)^{t-1}\big(J({\bm\theta^{*}})-J({\bm\theta_{0}})\big)+\frac{(M+1)(1-\gamma^{H})\zeta}{(1-\gamma)}}
\end{equation}
Thus,
\begin{equation}\label{vavarsg4}
E\bigg[\min_{{\bm p}\in \Delta}\sum_{t\in \mathcal{T}_m}f_t({\bm p})-\sum_{t\in \mathcal{T}_m}\min_{{\bm p}\in \Delta}f_t({\bm p})\bigg]\leq{\mathcal{O}(|\mathcal{B}|\sqrt{|\mathcal{B}|M})}
\end{equation}
Thus, lemm 2 is proven 
\section{Implementations}\label{impl}
All implementations are in python 3.6.8 using pytorch 0.4.0. 
\subsection{DDPG}
DDPG's agent uses actor-critic architecture. For the actor, we use three layers neural network with fully connected input layer mapping the states to a fully connected $400$ hidden layer followed by a fully connected $300$ output layer where its output size is equal to the action dimension. All input and hidden layers were followed by a rectifier nonlinearity while the output layer is a tanh layer to bound the actions. For the critic, we use neural network with three fully connected layers. All layers excluding the last one are followed by a rectifier nonlinearity. The first layer maps states to $400$ hidden layer. Actions are concatenated with the output of the first layer and fed to the second layer with $300$ output size. The third layer maps the $300$ ouput of the second layer to output of size $1$. Adam optimiser is used with its learning rates set to $10^{−4}$ and $10^{−3}$ for the actor and critic respectively. Mini-batch size is set to $64$. The rest hyper-parameter of DDPG are set the same as its original paper~\cite{lillicrap2015continuous}. 
\subsection{SAC}
SAC adopts the soft Q-learning (SQL) implementation of~\cite{haarnoja2017reinforcement} with two Q-functions. Both, the Q-value functions and policy / sampling network are neural networks comprised of $256$ hidden layer and ReLU nonlinearity. Adam optimiser is used with its learning rates set to $10^{−4}$, batch size is set to $256$. The rest hyper-parameter of SAC are set the same as in the orginal code~\footnote{https://github.com/vitchyr/rlkit}. 
\subsection{AES}
AES implementations resemble the steps presented in Alg.~\ref{alg2} with few practical variations. The AES hard resetting presented in Alg.~\ref{alg2} (line 9) is replaced with soft resetting version where forgetting factor $\varrho$ is used. For some experiments, we anneal this forgetting factor from initial to final values. For parameters tuning of AES, samples of different hyper-parameter settings are tested and the best ones are used. Table \ref{ParTab} lists the AES hyper-parameters used in the comparative evaluation in Fig.~\ref{fi4} and Fig.~\ref{fi5}. 

\begin{table*}[ht]
\caption{AES Hyper-parameters}
\label{ParTab}
\centering
\begin{tabular}{lll}\hline
 \toprule
Paramertes & values  \\\hline
AES-DDPG&&\\
\quad exploration rate $\kappa$ & $0.1$&\\
\quad regularisation factor $\nu$ &$1000$&\\
\quad forgetting factor $\varrho$ &$0.9$&\\
\hline
AES-SAC  &&\\
\quad Reacher, Walker2d, Halfcheetah  &&\\
\quad \quad exploration rate $\kappa$ & $0.2$&\\
\quad \quad regularisation factor $\nu$ &$1000$&\\
\quad \quad forgetting factor $\varrho$ &$0.7$&\\
\hline
\quad Ant, Hopper,   &&\\
\quad \quad exploration rate $\kappa$ & $0.2$&\\
\quad \quad regularisation factor $\nu$ &$1000$&\\
\quad \quad annealed forgetting factor $\varrho$ &$0.8\rightarrow 0.2$&\\
\hline
\quad InvertedDoublePendulumn, InvertedPendulumn($\nu=1000$), Humanoid ($\kappa=0.1$)   &&\\
\quad \quad exploration rate $\kappa$ & $0.2$&\\
\quad \quad regularisation factor $\nu$ &$10000$&\\
\quad \quad annealed forgetting factor $\varrho$&$0.7\rightarrow 0.2$&\\
 \bottomrule
\end{tabular}
\end{table*}  
\section{Algorithms}
\subsection{SAC application}\label{sec4}
We apply AES presented in Alg.~\ref{alg2} to SAC algorithm for continuous action proposed by~\cite{SAC_Haarnoja_2018}. We call the resulting algorithm AES-SAC presented in Alg.~\ref{alg23}.

\begin{algorithm}[t]
   \caption{Adaptive Experience selection for SAC: AES-SAC}\label{alg23}
\begin{algorithmic}[1]
\STATE \textbf{input:} number of iteration $T$, sampling batch $|\Psi|$, restarting period $M$, experience size $|\mathcal{B}|$, exploration meta-learning parameter $\kappa$, maximum trajectory length $H$
\STATE \textbf{initialise:} initialise policy parameters $\theta_p$, soft Q-function parameters $\theta_q$, value function parameters $\theta_v$ target value function parameters $\theta'_v$, probability of sampling weights $\bigg\{w(i)=0\bigg\}_{i=1}^{|\mathcal{B}|}$ and gradient accumulators $d\theta_p\leftarrow 0$, $d\theta_v\leftarrow 0$ and $d\theta_q\leftarrow 0$.
\REPEAT
\STATE Initialise a random process $\mathcal{N}$ for action exploration
\STATE Get initial stat $s_1$
\FOR{$i\in \{1,...H\}$}
\STATE Update sampling distribution $\bigg\{ p(j)=(1-\kappa)\frac{\sqrt{ w(j)+{|\mathcal{B}|}\big(\frac{1}{\beta}L(\sqrt{D}+HR)\big)^2/\kappa}}{\sum_{j=1}^{|\mathcal{B}|}\sqrt{ w(j)+{|\mathcal{B}|}\big(\frac{1}{\beta}L(\sqrt{D}+HR)\big)^2/\kappa}}+\kappa/{|\mathcal{B}|}\bigg\}_{j=1}^{|\mathcal{B}|}$
\STATE Select action $a_i=\mu_{\theta_p}(s_i)+\mathcal{N}$ according to the current policy and exploration noise
\STATE Execute action $a_i$ and receive reward $r_i$ and the next state $s_{i+1}$
\STATE Store transition $(s_i,a_i,r_i,s_{i+1})$ in the experience buffer by overwriting experience indexed by $j\sim \frac{1-p}{|\mathcal{B}|-1}$
\STATE Sample indices $\{j_1,...j_{|\Psi|}\}\sim p_t$  and use them to select batch of experiences from the experience buffer: $\bigg\{(s_j,a_j,r_j,s_{j+1})\bigg\}_{j\in \{j_1,... j_{|\Psi|}\}}$
\STATE Compute policy gradient $d\theta_p$, value function gradient $d\theta_v$ soft-Q function gradient $d\theta_p$ using the $|\Psi|$ sampled experiences $\bigg\{(s_j,a_j,r_j,s_{j+1})\bigg\}_{j\in \{j_1,... j_{|\Psi|}\}}$.
\STATE Use probability of sampling $p$ with importance sampling to correct the bias in $d\theta_p$, $d\theta_v$ and $d\theta_q$.
\STATE Update policy parameters $\theta_p $ using $d\theta_p$, value function parameter $\theta_v$ using $d\theta_v$ and soft-Q function parameters $\theta_q $ using $d\theta_q$.
\STATE Compute $\{\tilde{d}_{t}(j)\}_{j\in \{j_1,... j_{|\Psi|}\}}$ using $d\theta_p$, $d\theta_v$ and $d\theta_q$  and update $\bigg\{w(j)\leftarrow w(j)+\tilde{d}(j)/p(j)\bigg\}_{j\in \{j_1,... j_{|\Psi|}\}}$
\ENDFOR
\UNTIL{Max iteration $T$ reached}
\end{algorithmic}
\end{algorithm}
\subsection{DDPG application}\label{sec5}
We apply AES presented in Alg.~\ref{alg2} to DDPG algorithm for continuous action proposed by~\cite{lillicrap2015continuous}. We call the resulting algorithm AES-DDPG presented in Alg.~\ref{alg25}. 
\begin{algorithm}[t]
   \caption{Adaptive Experience selection for DDPG: AES-DDPG}\label{alg25}
\begin{algorithmic}[1]
\STATE \textbf{input:} number of iteration $T$, sampling batch $|\Psi|$, restarting period $M$, experience size $|\mathcal{B}|$, exploration meta-learning parameter $\kappa$, maximum trajectory length $H$
\STATE \textbf{initialise:} initialise policy parameters $\theta$, critic parameters $\theta_v$, target policy parameters $\theta'$, target critic parameters $\theta'_v$, probability of sampling weights $\bigg\{w(i)=0\bigg\}_{i=1}^E$ and gradient accumulators $d\theta\leftarrow 0$, $d\theta_v\leftarrow 0$.
\REPEAT
\STATE Initialise a random process $\mathcal{N}$ for action exploration
\STATE Get initial stat $s_1$
\FOR{$i\in \{1,...H\}$}
\STATE Update sampling distribution $\bigg\{ p(j)=(1-\kappa)\frac{\sqrt{ w(j)+{|\mathcal{B}|}\big(\frac{1}{\beta}L(\sqrt{D}+HR)\big)^2/\kappa}}{\sum_{j=1}^{|\mathcal{B}|}\sqrt{ w(j)+{|\mathcal{B}|}\big(\frac{1}{\beta}L(\sqrt{D}+HR)\big)^2/\kappa}}+\kappa/{|\mathcal{B}|}\bigg\}_{j=1}^{|\mathcal{B}|}$
\STATE Select action $a_i=\mu_{\theta}(s_i)+\mathcal{N}$ according to the current policy and exploration noise
\STATE Execute action $a_i$ and receive reward $r_i$ and the next state $s_{i+1}$
\STATE Store transition $(s_i,a_i,r_i,s_{i+1})$ in the experience buffer by overwriting experience indexed by $j\sim \frac{1-\boldsymbol p}{|\mathcal{B}|-1}$
\STATE Sample indices $\{j_1,...j_{|\Psi|}\}\sim p_t$  and use them to select batch of experiences from the experience buffer: $\bigg\{(s_j,a_j,r_j,s_{j+1})\bigg\}_{j\in \{j_1,... j_{|\Psi|}\}}$
\STATE Compute critic gradient $d\theta_v$ and actor gradient $d\theta$ using the $|\Psi|$ sampled experiences $\bigg\{(s_j,a_j,r_j,s_{j+1})\bigg\}_{j\in \{j_1,... j_{|\Psi|}\}}$
\STATE Use probability of sampling $p$ with importance sampling to correct the bias in $d\theta_v$ and $d\theta$.
\STATE Update critic parameters $\theta_v $ using $d\theta_v$ and actor parameters $\theta$ using $d\theta$
\STATE Compute $\{\tilde{d}_{t}(j)\}_{j\in \{j_1,... j_{|\Psi|}\}}$ using $d\theta$ and $d\theta_v$  and update $\bigg\{w(j)\leftarrow w(j)+\tilde{d}(j)/p(j)\bigg\}_{j\in \{j_1,... j_{|\Psi|}\}}$
\ENDFOR
\UNTIL{Max iteration $T$ reached}
\end{algorithmic}
\end{algorithm}
\section{Additional results}
\subsection{Variance evaluation}\label{variastud}
\begin{figure}[t]
\centering
\includegraphics[width=\textwidth,height=8.7cm]{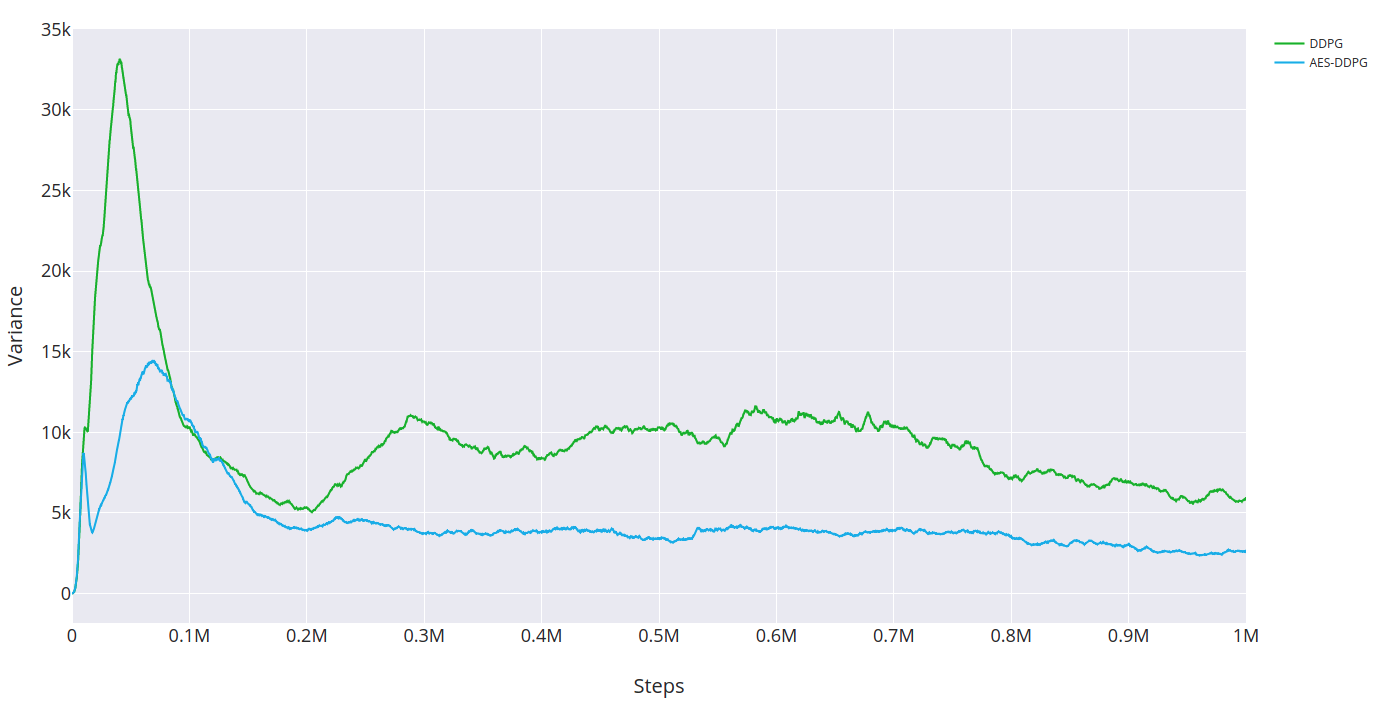}
\caption{Variance on Walker2d}
\label{ffg}
\end{figure}

Variance of DDPG on Walker2d environment is shown in Fig.~\ref{ffg}. We can clearly notice the improvement by AES-DDPG compared to DDPG on walker2d which is considered as complex environment.
\subsection{Numerical measurements}\label{nummea}
When applying AES, we are interested in the improvement in sample efficiency, that is, can we achieve higher score using same amount of samples. We are also interested in the \textit{learning stability} and \textit{final performance} because reducing variance should affect these measurements. To report the effect of AES on this three aspect, we define \textit{Learning speed} as the average speed with respect to steps within the steps needed to reach the maximum score during the last 60\% of total learning steps; \textit{Learning speed}= (\textit{max score})/(\textit{number of steps}). Since our method reduce the updating variance by selecting samples with less noisy gradient, faster improvement can be achieved. That is because local optimum can be reached without much of distraction, allowing better optimum using less samples. That reflect higher score with less steps. Thus, we also report the (\textit{max score}) during the last 60\% of total learning episodes. Note that the score denotes the per-episode total testing return and is computed using moving average windows. 

A common drawback of DRL algorithms is that even when a good performance has been achieved, it can drop significantly as the distribution of acquired data changes. That happens when a reached local optimum is lost to a worse one. Hence, the algorithm de-learns and could diverge. By using our variance reduction methods, distracting samples are avoided. Thus, the algorithm is expected to leave a optimum only to reach a better one. To measure this \textit{Learning stability}, we compute the proportion of the mean scores achieved at the end of learning to the max score achieved. This expresses how much of the \textit{max score} has been carried till the end of the learning. We also report the \textit{final performance} which is the testing score after learning is over. The average and standard deviation score is computed over 5 different seeds $\{2,20,200,2000,20000\}$. We report the results \textit{robustness} which is expressed by the mean of the standard deviation over the last 20\% steps. 

Table~\ref{acer} shows the numerical figures of the measurements discussed above: \textit{learning speed}, \textit{learning stability}, \textit{Max score}, \textit{Robustness} and \textit{Final performance}. These results summarise the learning performance reported in Fig.~\ref{fi4} and Fig.~\ref{fi5}. 

\begin{table*}[]
\centering
\caption{Performance of AES applied to DDPG and SAC in terms of sample efficiency, learning stability and final performance}\label{acer}
\begin{tabular}{|l|l|l|l|l|l|l|l|}
\hline
Environment               & Algorithm             & Method & Learning speed & Learning stability & Max score & Robustness & Final performance \\ \hline
Humanoid & DDPG & Fifo   &    $0.00169$            &    $1$                &   $287.537$        & $103.397$           & $287.537$                  \\ \cline{3-8} 
                          &                       & PriExp &    $0.00253$            &    $0.999$                &   $382.384$        & $\bold{81.240}$           & $382.254$      \\ \cline{3-8} 
                          &                       & AES    &     $\bold{0.00259}$           &      $1$              &  $\bold{391.208}$         &   $112.204$         &       $\bold{391.208}$             \\ \cline{2-8} 
                          & SAC  & Fifo   &  $0.00183$              &   $0.999$                 &   $5500.471$        &    $\bold{159.036}$        &   $5500.234$                \\ \cline{3-8} 
                          &                       & PriExp &    $\bold{0.00189}$            &     $\bold{1}$               &  $\bold{5692.541}$         &   $288.715$         &  $\bold{5692.541}$                 \\ \cline{3-8} 
                          &                       & AES    &    $0.00187$            &  $0.999$                  &   $5612.325$        &  $242.891$          &  $5607.831$                 \\ \hline
Ant & DDPG & Fifo   &  $0.0016$              &  $0.797$                  &    $\bold{890.913}$       &    $121.338$        &    $710.784$               \\ \cline{3-8} 
                          &                       & PriExp &  $0.0012$              & $\bold{0.984}$                  &  $873.657$         &   $51.073$         &     $860.445$              \\ \cline{3-8} 
                          &                       & AES    &  $\bold{0.0019}$              &   $0.981$                 &  $887.672$         &    $\bold{35.888}$        &  $\bold{871.596}$                 \\ \cline{2-8} 
                      & SAC  & Fifo   &  $0.0018$              &  $0.942$                  &   $5239.381$        &    $1018.327$        &      $4936.972$             \\ \cline{3-8} 
                          &                       & PriExp & $0.00171$               &  $0.977$                  &   $4523.774$        &   $734.194$         &  $4422.798$                 \\ \cline{3-8} 
                          &                       & AES    &   $\bold{0.002}$             &  $\bold{0.999}$                  &    $\bold{5982.045}$       &  $\bold{299.794}$          &     $\bold{5977.75}$              \\ \hline
Walker2d & DDPG & Fifo   & $0.0016$               &  $0.925$                  & $1344.017$          &  $541.343$          &  $1244.302$                 \\ \cline{3-8} 
                          &                       & PriExp &    $0.0019$            & $1$                   & $1927.011$          &      $329.567$      &         $1927.011$          \\ \cline{3-8} 
                          &                       & AES    & $\bold{0.0026}$               &   $1$                 &   $\bold{2654.077}$        &    $\bold{309.412}$        &       $\bold{2654.077}$             \\ \cline{2-8} 
                          & SAC  & Fifo   &   $0.0038$             &      $1$              & $3818.676$          &   $507.097$         &     $3818.676$              \\ \cline{3-8} 
                          &                       & PriExp &      $0.00302$          & $1$                   &   $3025.566$        &     $810.013$       &  $3025.566$                 \\ \cline{3-8} 
                          &                       & AES    &   $\bold{0.00439}$             &   $1$                 &    $\bold{4389.851}$       &   $\bold{326.548}$         &   $\bold{4389.851}$                \\ \hline
Hopper & DDPG & Fifo   & $0.0025$               &  $0.854$                  &   $1958.503$        & $952.031$           &    $1674.366$               \\ \cline{3-8} 
                          &                       & PriExp &  $\bold{0.0035}$              & $0.928$                   &    $2400.457$       &    $304.08$        &     $2229.206$              \\ \cline{3-8} 
                          &                       & AES    & $0.0026$                &   $\bold{1}$                 &      $\bold{2600.677}$     & $\bold{281.619}$           & $\bold{2600.677}$                  \\ \cline{2-8} 
                          & SAC  & Fifo   & $0.0027$               &   $0.912$                 &  $2643.635$         &   $\bold{545.558}$         &     $2413.141$              \\ \cline{3-8} 
                          &                       & PriExp &    $0.0027$            &   $0.999$                 &  $2649.586$         &    $588.046$        &  $2648.707$                 \\ \cline{3-8} 
                          &                       & AES    &  $\bold{0.0029}$              &   $\bold{1}$                 &  $\bold{2947.33}$         &    $564.625$        &     $\bold{2947.33}$              \\ \hline
HalfCheetah & DDPG & Fifo   &  $0.0064$              &    $1$                & $6429.229$          &  $913.628$          &  $6429.229$                 \\ \cline{3-8} 
                          &                       & PriExp & $0.0068$               &    $1$                &  $6850.154$         &  $459.511$         &  $6850.154$                 \\ \cline{3-8} 
                          &                       & AES    &  $\bold{0.0075}$              &   $1$                 &    $\bold{7533.83}$       &  $\bold{457.035}$          &   $\bold{7533.837}$                \\ \cline{2-8} 
                          & SAC  & Fifo   &    $0.0033$            &     $1$               &   $9944.207$        &  $3793.857$         &     $9944.207$              \\ \cline{3-8} 
                          &                       & PriExp &      $0.0029$          &    $1$                &   $8712.836$        &   $4376.23$         &     $8712.836$              \\ \cline{3-8} 
                          &                       & AES    &     $\bold{0.0039}$           &    $1$                &  $\bold{11792.044}$         &   $\bold{2640.481}$         &   $\bold{11792.044}$                \\ \hline
InvertedDoublePendulum & DDPG & Fifo   & $\bold{0.02033}$               &   $0.789$                 &      $8214.756$     &    $1158.533$        &    $6481.968$               \\ \cline{3-8} 
                          &                       & PriExp &  $0.0085$              &    $0.999$                &   $8545.812$        &    $714.948$        &  $8545.053$                 \\ \cline{3-8} 
                          &                       & AES    &   $0.0088$             &  $0.999$                  &  $\bold{8761.786}$         &  $\bold{459.784}$          &   $\bold{8754.715}$                \\ \cline{2-8} 
                          & SAC  & Fifo   &   $\bold{0.062}$             &    $0.995$                &   $9358.604$        &   $495.93$         &     $9319.816$              \\ \cline{3-8} 
                          &                       & PriExp &   $0.041$             &  $0.995$                  &  $\bold{9358.608}$         &  $218.671$          &   $9320.195$                \\ \cline{3-8} 
                          &                       & AES    &   $0.043$             & $\bold{0.999}$                   & $9358.223$          &    $\bold{48.287}$        &     $\bold{9356.588}$              \\ \hline
InvertedPendulum & DDPG & Fifo   &  $\bold{0.0024}$              &  $0.927$                  &  $\bold{982.75}$         &    $\bold{31.326}$        &     $911.492$              \\ \cline{3-8} 
                          &                       & PriExp &     $0.0022$           &  $0.941$                 &  $977.957$         &     $54.590$       &     $920.393$              \\ \cline{3-8} 
                          &                       & AES    &   $0.0012$             &    $\bold{0.988}$                &    $972.923$       &  $32.544$          &       $\bold{961.764}$            \\ \cline{2-8} 
                          & SAC  & Fifo   &   $0.0065$             &    $0.928$                &   $1000$        &   $29.824$         &     $928.559$ \\ \cline{3-8} 
                          &                       & PriExp & $0.0035$               &    $1$                &     $0.0035$      &      $187.152$      &   $1000$                \\ \cline{3-8} 
                          &                     & AES    &    $\bold{0.0082}$            &    $1$                &      $1000$     &    $\bold{4.943}$        &    $1000$               \\ \hline
Reacher & DDPG & Fifo   &$-1.531e^{-5}$                & $0.918$                   &  $-10.028$         &   $1.572$         &   $-10.922$                \\ \cline{3-8} 
                          &                       & PriExp &  $-2.382e^{-5}$              &  $0.896$                  &     $-10.365$      &  $1.544$          &  $-11.566$                 \\ \cline{3-8} 
                          &                       & AES    &  $\bold{-9.68e^{-6}}$              &$\bold{1}$                    &  $\bold{-9.680}$         &  $\bold{0.937}$          & $\bold{-9.68}$                  \\ \cline{2-8} 
                          & SAC  & Fifo   &  $-5.366$              & $0.982$                   &   $\bold{-0.214}$        & $0.026$           &    $-0.218$               \\ \cline{3-8} 
                          &                       & PriExp &    $-6.335$            & $0.965$                   &   $-0.253$        &  $0.037$          &   $-0.262$                \\ \cline{3-8} 
                          &                       & AES    &  $\bold{-2.391}$              &    $\bold{0.998}$                &    $-0.217$       &  $\bold{0.019}$          &   $\bold{-0.217}$                \\ \hline
\end{tabular}
\end{table*}
\end{document}